\documentclass{article}

\usepackage{arxiv}

\usepackage[utf8]{inputenc} 
\usepackage[T1]{fontenc}    
\usepackage{hyperref}       
\usepackage{url}            
\usepackage{booktabs}       
\usepackage{amsfonts}       
\usepackage{nicefrac}       
\usepackage{microtype}      
\usepackage{lipsum}		
\usepackage{graphicx}
\usepackage{doi}
\usepackage{amsthm}
\usepackage{amsmath}
\theoremstyle{plain}

\theoremstyle{definition}
\newtheorem{definition}{Definition}

\theoremstyle{remark}
\newtheorem{remark}{Remark}

\title{Clustering in hyperbolic balls}

\author{Vladimir Ja\' cimovi\' c \\
	Faculty of Natural Sciences and Mathematics\\
	University of Montenegro\\
	Cetinjski put bb., 81000 Podgorica\\
	Montenegro\\
	\texttt{vladimirj@ucg.ac.me} \\
        \And
	Aladin Crnki\' c \\
	Faculty of Technical Engineering\\
	University of Biha\' c\\
	Irfana Ljubijanki\' ca bb., 77000 Biha\' c\\
	Bosnia and Herzegovina\\
	\texttt{aladin.crnkic@unbi.ba} \\
}

\renewcommand{\shorttitle}{Clustering in hyperbolic balls}

\hypersetup{
pdftitle={Clustering in hyperbolic balls},
pdfsubject={-},
pdfauthor={Vladimir Ja\' cimovi\' c, Aladin Crnki\' c},
pdfkeywords={??,??},
}

\begin{document}
\maketitle

\begin{abstract}
The idea of representations of the data in negatively curved manifolds recently attracted a lot of attention and gave a rise to the new research direction named {\it hyperbolic machine learning} (ML). In order to unveil the full potential of this new paradigm, efficient techniques for data analysis and statistical modeling in hyperbolic spaces are necessary. In the present paper rigorous mathematical framework for clustering in hyperbolic spaces is established. First, we introduce the $k$-means clustering in hyperbolic balls, based on the novel definition of barycenter. Second, we present the expectation-maximization (EM) algorithm for learning mixtures of novel probability distributions in hyperbolic balls. In such a way we lay the foundation of unsupervised learning in hyperbolic spaces.
\end{abstract}

\keywords{$k$-means \and expectation-maximization \and M\" obius transformation \and hyperbolic machine learning}

\section{Introduction}\label{sec:1}
Learning low-dimensional representations of the data is one of central issues in modern ML. This issue became particularly relevant with the advent of deep learning architectures with latent spaces. This motivated investigations of compact models with curved latent spaces, thus transcending beyond the traditional Euclidean paradigm in ML \cite{Smith,Sala,Arvanitidis}. Embedding the data into hyperbolic spaces seems like an advantageous approach to a wide range of tasks. The underlying idea is that important structural information in certain ubiquitous data sets is preserved in manifolds with the negative curvature. Most notably, there are strong arguments that hierarchies, analogies and hypernymies are naturally represented in hyperbolic spaces.

Nowadays, the potential of hyperbolic ML is widely recognized with experiments in various fields, including
natural languages \cite{Nickel,Tifrea}, networks \cite{Muscoloni,Chami}, biological data \cite{Corso,Facebook}, user preferences \cite{Li,Chamberlain}, brain \cite{Baker,Allard} and many more.

Despite very encouraging results, the lack of principled approaches and rigorous mathematical techniques imposes severe limitations on future advances in hyperbolic ML. Further advances rely on systematization and extension of the mathematical framework for statistical modeling and optimization in hyperbolic spaces.

Hyperbolic representations can be advantageous for those kinds of data that are difficult to quantify, but exhibit some hierarchical structure (such as words, molecules, preferences, sentiments, etc.). Therefore, the first stage of any hyperbolic ML solution must be an algorithm for embedding the data. These embedding algorithms are almost always demanding and are important and inevitable components of (future) hyperbolic DL pipelines. Most of research efforts have focused on this aspect of hyperbolic ML. Nowadays, methods for embedding the data into hyperbolic spaces are fairly well developed, although many challenges remain, mostly due to the difficulty of estimation and optimization in hyperbolic spaces.

Once the data are embedded in hyperbolic spaces, the second stage relies on tools for data analysis and generative modeling. Only with further improvements of existing embedding algorithms, as well as the development of (un)supervised learning methods, the sufficient machinery for the design of hyperbolic ML pipelines will be available. In particular, clustering is the cornerstone of unsupervised learning. In the present paper we address this issue.


Emphasize that clustering in Euclidean spaces is very well established, based on the clear and universally accepted underlying mathematical apparatus. Indeed, notion of the mean (average) in Euclidean spaces is obvious. Moreover, the computational procedure for averaging vectors is very easy. When it comes to statistics, Gaussian probability distributions serve as almost universal model for encoding uncertainties. Since the representative power of the Gaussian family is very modest, mixtures of Gaussians are commonly used for generative modeling and inference. There are several methods for training the Gaussian mixture models, among which the expectation-maximization (EM) algorithm is the most popular. More generally, this famous algorithm \cite{Dempster} can be applied to various problems with latent variables and plays the central role in generative and statistical modeling.

When it comes to non-Euclidean data, clustering algorithms on spheres are less known, but reasonably well established. EM algorithms for learning mixtures of von Mises-Fisher distributions \cite{Banerjee} and Kent distributions \cite{Hamelryck} for spherical data are proposed. The present study is motivated by the lack of analogous clustering methods for hyperbolic data.

The exposition is organized along the following lines.

In the next Section we provide a very brief overview of basic facts about Poincar\' e balls and their groups of isometries. In sections \ref{sec:3} and \ref{sec:4} we introduce the mathematical apparatus which is necessary for precise clustering. The notion of (weighted) barycenter in hyperbolic ball, as well as the corresponding computational procedure are presented in Section \ref{sec:3}. In Section \ref{sec:4} we introduce the family of probability distributions and present techniques for sampling of random points in hyperbolic balls, as well as maximum likelihood algorithm for estimation of parameters. In Section \ref{sec:5} we propose the algorithms for the minimal model of hyperbolic balls - two-dimensional manifold, well-known as the Poincar\' e disc. The methods are validated in three experiments with the data in the Poincar\' e disc. In Section \ref{sec:6} we extend algorithms to Poincar\' e balls of an arbitrary dimension. Experimental results on two data sets are reported for the 3-dimensional hyperbolic ball. Finally, Section \ref{sec:7} contains concluding remarks and the discussion on significance of the presented methods.

\section{Preliminaries: Poincar\' e balls and their symmetry groups}\label{sec:2}

In the present Section we introduce some basic facts about Poincar\' e balls and their isometries.

Denote by $\langle x,y \rangle = x_1y_1 + \cdots + x_n y_n$ the scalar product in $\mathbb{R}^n$ and $|x| = \sqrt{\langle x,x \rangle}$.

\begin{definition}
\label{Poin_ball_def}
Consider the set $\{x \in \mathbb{R}^n \; : \; |x|<1\}$ equipped with the metric tensor
$$
ds^2 = 4 \frac{d x_1^2 + \cdots + d x_n^2}{(1-|x|^2)^2}.
$$
We denote this manifold by $\mathbb{B}^n$ and name it the $n$-dimensional Poincar\' e ball.
\end{definition}

Hyperbolic measure in $\mathbb{B}^n$ is defined as
\begin{equation}
\label{hyp_meas_ball}
d \Lambda(x) = \frac{d \lambda(x)}{(1-|x|^2)^n}.
\end{equation}
where $d \lambda(x)$ denotes the Lebesgue measure in $\mathbb{R}^n$.

Let $x,y \in \mathbb{B}^n$. The following formula for the Poincar\'e distance in $\mathbb{B}^n$ holds
\begin{equation}\label{pome}d_{hyp}(x,y)=\frac{1}{2}\log \frac{1+R}{1-R},\end{equation} where
 \begin{equation}
 \label{rho}
 R=\frac{|x-y|}{\sqrt{\rho(x,y)}} \mbox{ and } \rho(x,a)=|x-a|^2+(1-|a|^2)(1-|x|^2).
 \end{equation}

Isometries of $\mathbb{B}^n$ are given by the following formula
\begin{equation} \label{Mobius_ball}
h_a(x)= A \frac{a|x-a|^2+(1-|a|^2)(a-x)}{\rho(x,a)},
\end{equation}
where $A$ is the orthogonal transformation of the Euclidean space $\mathbb{R}^n$.

Mappings of the form \eqref{Mobius_ball} are named {\it M\" obius transformations} of the unit ball.

We denote the group of M\" obius transformations by ${\mathbb G}_n$. Notice that $\mathbb{G}_n$ is isomorphic to the Lorentz group $SO^+(n,1)$.

Notice that we have chosen parametrization of M\" obius transformations \eqref{Mobius_ball} in such a way to have $h_c^{-1}(x)=h_c(x)$ for every $c\in \mathbb{B}^n$.

Group ${\mathbb G}_n$ operates on the set ${\cal P}(\mathbb{B}^n)$ of all probability measures on $\mathbb{B}^n$. Given a measure $\mu \in {\cal P}(\mathbb{B}^n)$ we will use the notation $g_* \mu$ for the pullback measure defined as
\begin{equation}
\label{pull_back}
g_* \mu (S) = \mu(g^{-1}(S)), \mbox{ for any Borel set } S \subseteq \mathbb{B}^n.
\end{equation}





\subsection{The Poincar\' e disc}

For the particular case $n=2$ Definition \ref{Poin_ball_def} yields the two-dimensional model of hyperbolic geometry named Poincar\' e disc. In this dimension it is convenient to use the algebra of complex numbers and the framework of complex analysis.

Consider the open disc in the complex plane $\{ z \in \mathbb{C}: \; \; |z|<1 \}$ and introduce the distance by the following formula:
$$
g(w,z) = \log \left( \frac{|1-\bar w z| + |z-w|}{|1-\bar w z|-|z-w|} \right)
$$
for all complex numbers $z$ and $w$ satisfying $|z|<1$ and $|w|<1$.

\begin{definition}
The Riemannian manifold $\{ z \in \mathbb{C}: \; \; |z|<1 \}$ with the distance $g$ is called {\it the Poincar\' e disc}.
\end{definition}

We will denote the Poincar\' e disc by $\mathbb{B}^2$.

Isometries of $\mathbb{B}^2$ are M\" obius transformations acting on the complex plane of the following form
\begin{equation}
\label{Mobius}
g_a(z) = e^{i \theta} \frac{a-z}{1-\bar a z}, \quad \theta \in [0,2 \pi), \; a \in \mathbb{B}^2.
\end{equation}

Transformations of the form \eqref{Mobius} constitute a subgroup of the group of all M\" obius transformations (linear-fractional transformations) acting on the complex plane. We denote this (sub)group by $\mathbb{G}_2$.

\begin{remark}
The (sub)group $\mathbb{G}_2$ is isomorphic to the Lie group $SU(1,1)$ of matrices of the form
$$
\left(
\begin{array}{cc}
a & b \\
- \bar b & \bar a
\end{array}
\right),
 \mbox{  where  } a,b \in \mathbb{C}, \quad |a|^2 + |b|^2 = 1.
$$
More precisely, $\mathbb{G}_2$ is isomorphic to the quotient group $PSU(1,1) = SU(1,1) / \pm I$.
\end{remark}

The hyperbolic measure in $\mathbb{B}^2$ reads
\begin{equation}
\label{hyp_meas_disc}
d \Lambda(z) = \frac{d \lambda(z)}{(1-|z|^2)^2}.
\end{equation}

The next two sections contain concepts and techniques in Poincar\' e balls which are necessary for clustering algorithms.

\section{Barycenters in Poincar\' e balls}\label{sec:3}

In the present Section we answer the following two questions:

\begin{enumerate}
\item[i)] Given observations $y_1,\dots,y_N$ in $\mathbb{B}^n$ with the corresponding weights $w_1,\dots,w_N$, what is their weighted mean?

\item[ii)] How can we compute the (weighted) mean?

\end{enumerate}

\subsection{Conformal barycenters in Poincar\' e balls}

Let $\{y_1,\dots,y_N\}$ be points in the Poincar\' e e ball $\mathbb{B}^n$. Consider the following function
\begin{equation}
\label{potential_Poin_ball}
H_n(a) = - \sum_{i=1}^N \log \frac{(1-|a|^2)(1-|y_i|^2)}{\rho(y_i,a)}, \quad a \in \mathbb{B}^n,
\end{equation}
where $\rho(u,v)$ is defined in \eqref{rho}.

It has been shown in \cite{JK} that \eqref{potential_Poin_ball} is geodesically convex and, hence, has a unique minimizer in $\mathbb{B}^n$.

\begin{definition} \cite{JK}
Conformal barycenter of points $\{y_1,\dots,y_N\}$ in $\mathbb{B}^n$ is the unique minimizer of the function \eqref{potential_Poin_ball}.
\end{definition}

The conformal barycenter is conformal invariant. This means that if $a$ is the conformal barycenter of points $\{y_1,\dots,y_N \}$ and $h \in \mathbb{G}_n$, then $h(a)$ is the conformal barycenter of points $\{h(y_1),\dots,h(y_N) \}$.

\begin{figure*}[t]
\centering
  \begin{tabular}{@{}ccc@{}}
    \includegraphics[width=.3\textwidth]{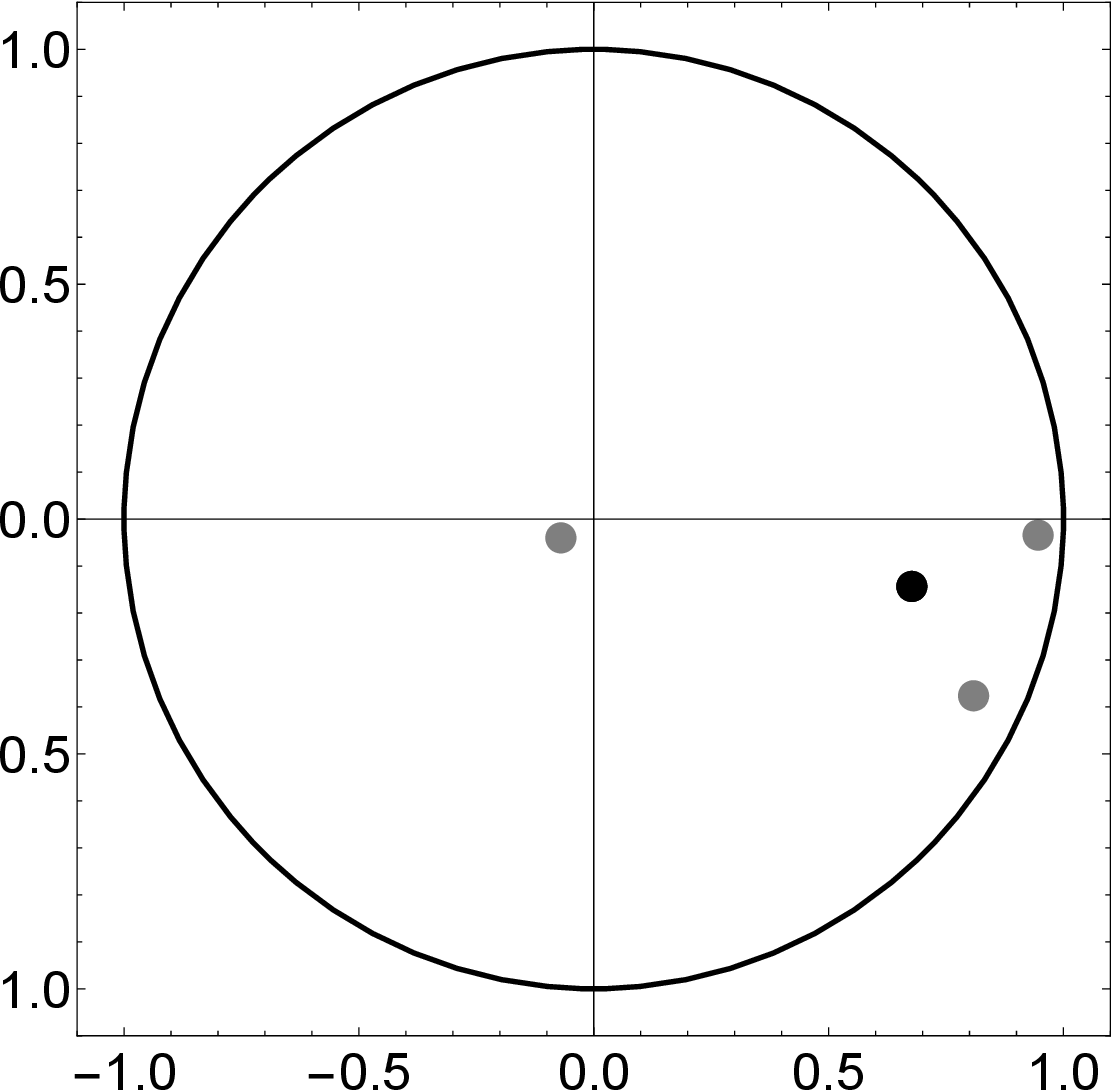}&\includegraphics[width=.3\textwidth]{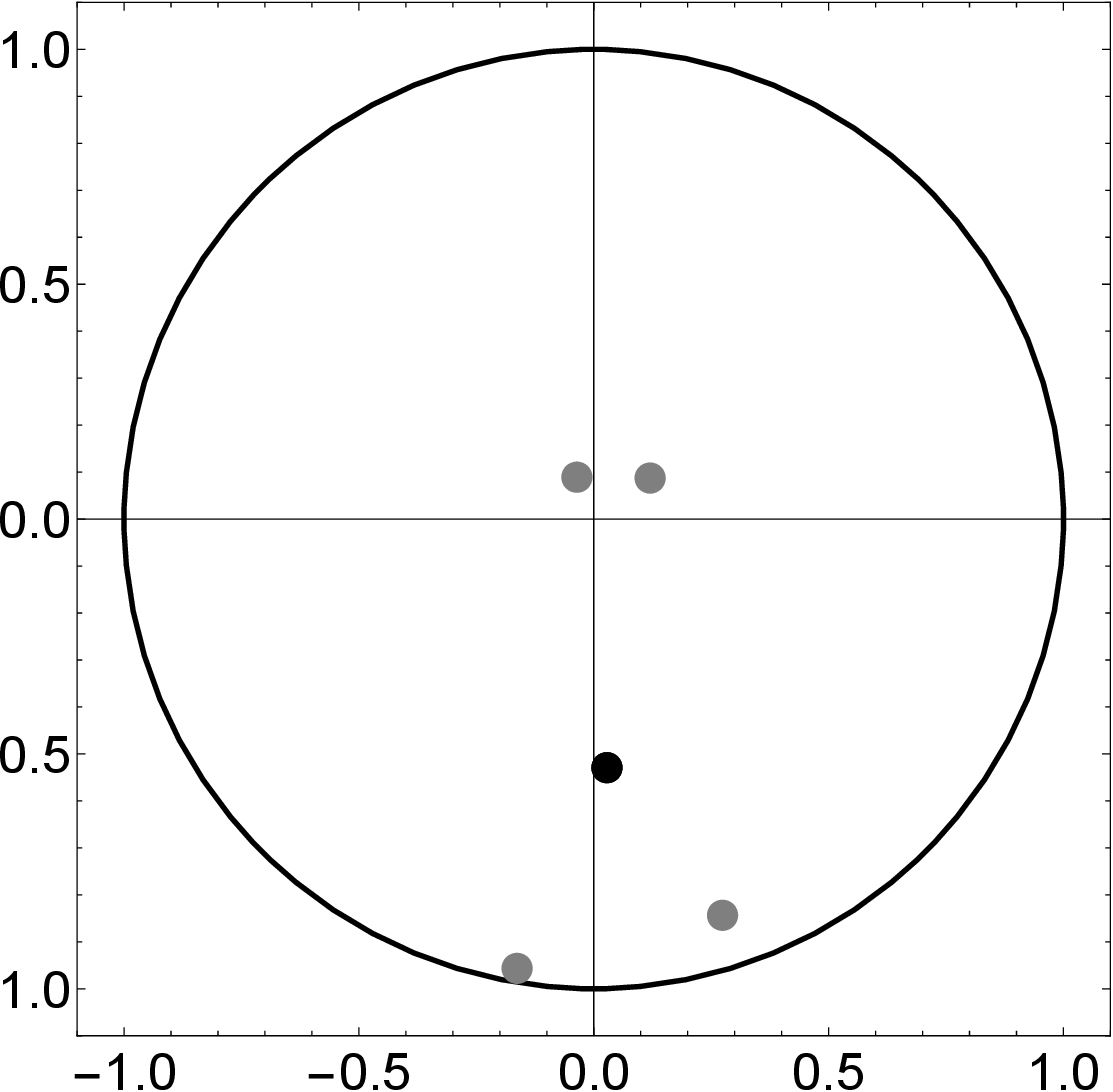}&\includegraphics[width=.3\textwidth]{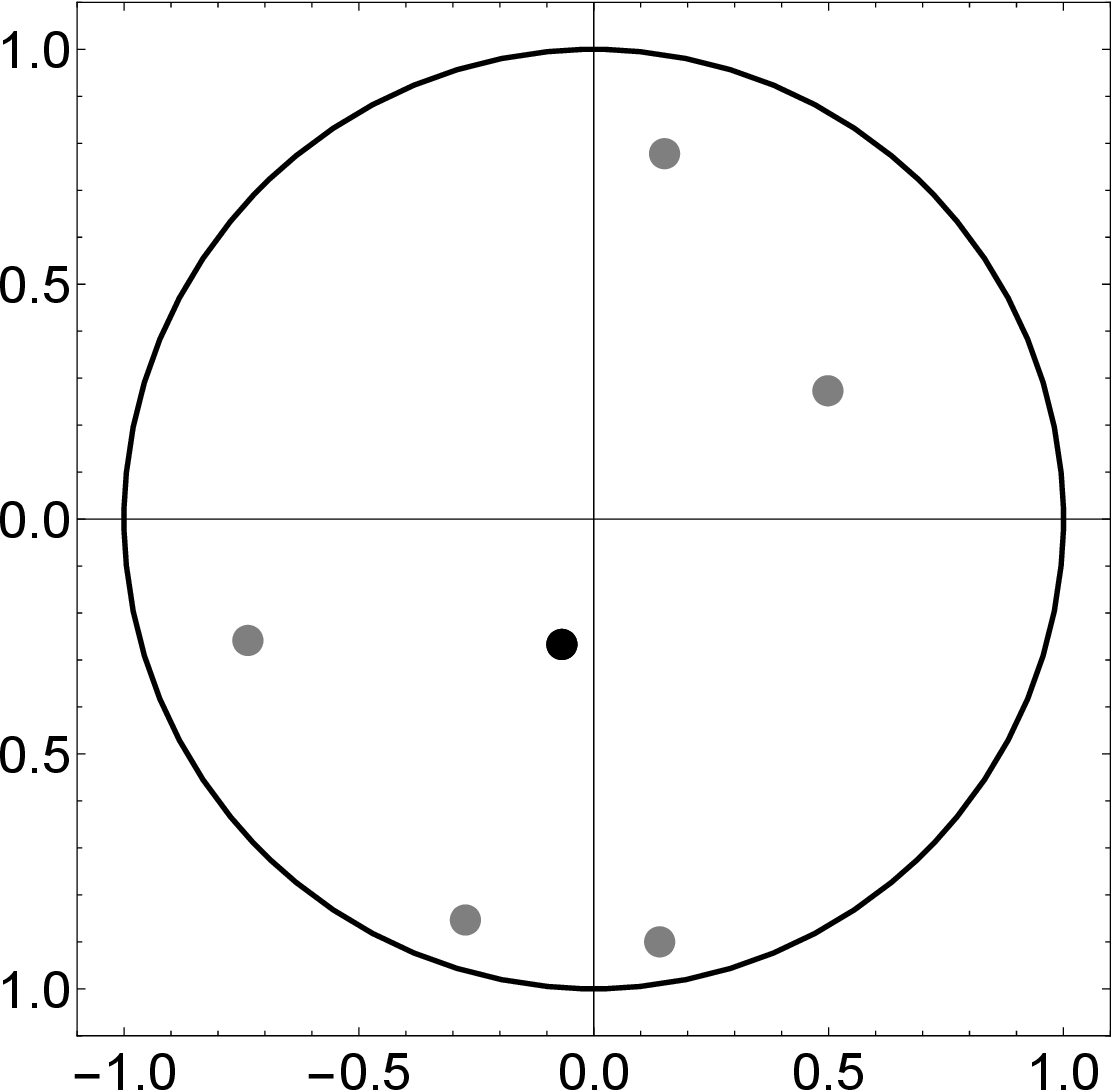}
  \end{tabular}
  \caption{\label{fig:1}Barycenters of configurations of points in the Poincar\' e disc}
\end{figure*}

We refer to \cite{JK} for rigorous proofs and more details about the conformal barycenter. In Figure \ref{fig:1} we plot conformal barycenters for three configurations of points.

The method for computation of the conformal barycenter has been proposed in \cite{Jacimovic}. Consider the following system of ODE's in $\mathbb{B}^n$
\begin{equation}
\label{swarm}
\dot x_i = \frac{1}{2}\left(1+\left|x_i\right|^{2}\right) - \langle x_i,f \rangle x_i, \quad i=1,\dots,N,
\end{equation}
where $f : \mathbb{R}^n \times \cdots \times \mathbb{R}^n \to \mathbb{R}^n$ is given by
\begin{equation}
\label{coupling}
f = \frac{K}{N}\sum\limits_{j=1}^{N}x_j \mbox{    with  } K<0.
\end{equation}

The dynamics \eqref{swarm} preserves the unit ball, meaning that if initial conditions $x_1(0),\dots,x_N(0)$ belong to $\mathbb{B}^n$, then solutions of \eqref{swarm} $x_1(t),\dots,x_N(t)$ belong to $\mathbb{B}^n$ for any $t>0$.

Moreover, it has been shown in \cite{Jacimovic} that the system \eqref{swarm}-\eqref{coupling} is the gradient flow in hyperbolic metric for the potential function \eqref{potential_Poin_ball}.
Therefore, conformal barycenter of a set of points can be computed using the following gradient descent algorithm.

\begin{enumerate}
\item[i)] Let $y_1,\dots,y_N$ be points in the Poincar\' e ball ($N>2$).

\item[ii)] Solve (\ref{swarm}) with $f$ defined by \eqref{coupling} for initial conditions $x_1(0)=y_1,\dots,x_n(0)=y_N$.

\item[iii)] For a sufficiently large $T$, it holds that $x_1(T)+\cdots+x_N(T)=0$. This means (see \cite[Proposition 3]{Jacimovic}) that zero (center of the ball) is the conformal barycenter of points $x_1(T),\dots,x_N(T)$. Furthermore, due to \cite[Theorem 12]{Jacimovic}, there exists a (unique up to an orthogonal transformation) M\" obius transformation $h \in \mathbb{G}_n$, such that $x_1(T)=h(y_1),\dots,x_N(T)=h(y_N).$

\item[iv)] The point $a=h^{-1}(0)$ is the conformal barycenter of points $\{ y_1,\dots,y_N\}$.
\end{enumerate}

\subsection{Weighted barycenters in Poincar\' e balls}
\label{weight_bary}

Let $\{y_1,\dots,y_N\}$ be observations in $\mathbb{B}^n$ with the corresponding weights $w_1>0,\dots,w_N>0$. Define the following function
\begin{equation}
\label{weight_potential_Poin_ball}
\tilde H_n(a) = - \sum_{i=1}^N w_i \log \frac{(1-|a|^2)(1-|y_i|^2)}{\rho(y_i,a)},
\end{equation}
where $\rho(u,v)$ is defined in \eqref{rho}.

Using the fact that \eqref{potential_Poin_ball} is geodesically convex, it is straightforward to show that \eqref{weight_potential_Poin_ball} with non-negative weights $\{w_i\}$ has a unique minimum in $\mathbb{B}^n$.

\begin{definition}
Weighted barycenter of points $y_1,\dots,y_N$ with the weights $w_1,\dots,w_N$ is the unique minimum of the function \eqref{potential_Poin_ball}.
\end{definition}

\begin{figure*}[t]
\centering
  \begin{tabular}{@{}ccc@{}}
    \includegraphics[width=.3\textwidth]{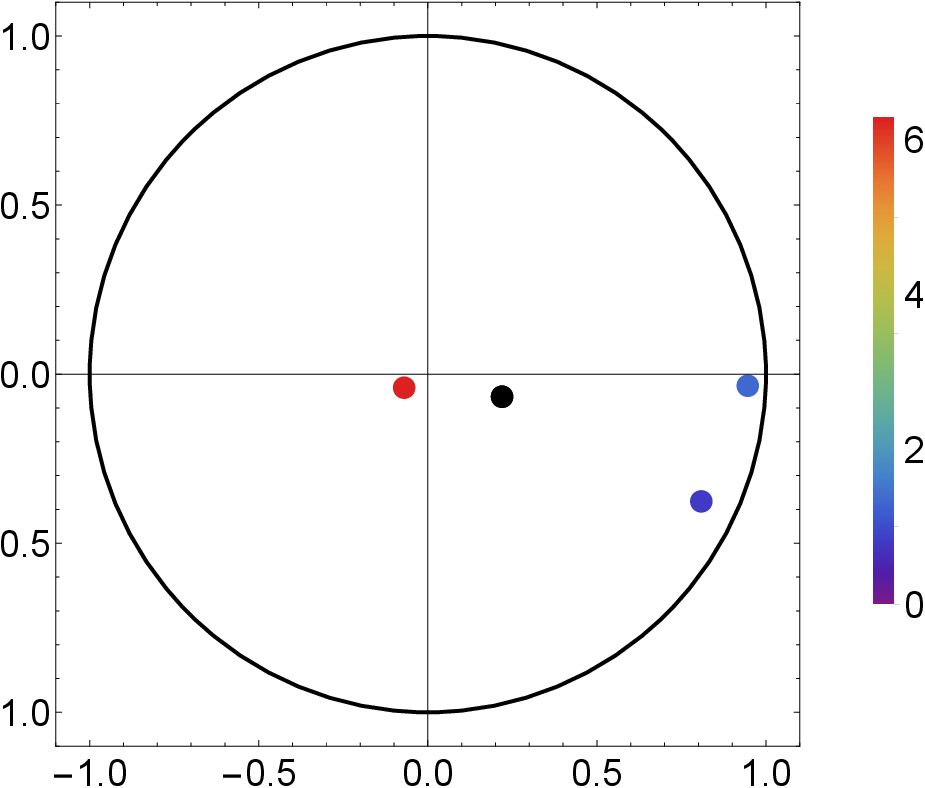}&\includegraphics[width=.3\textwidth]{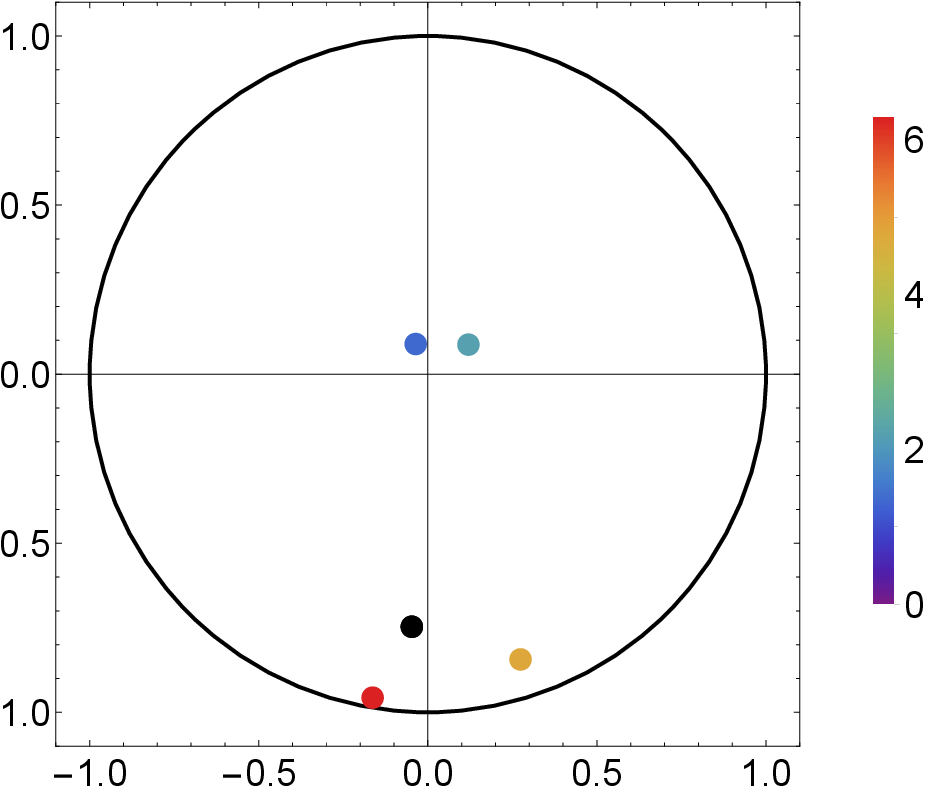}&\includegraphics[width=.3\textwidth]{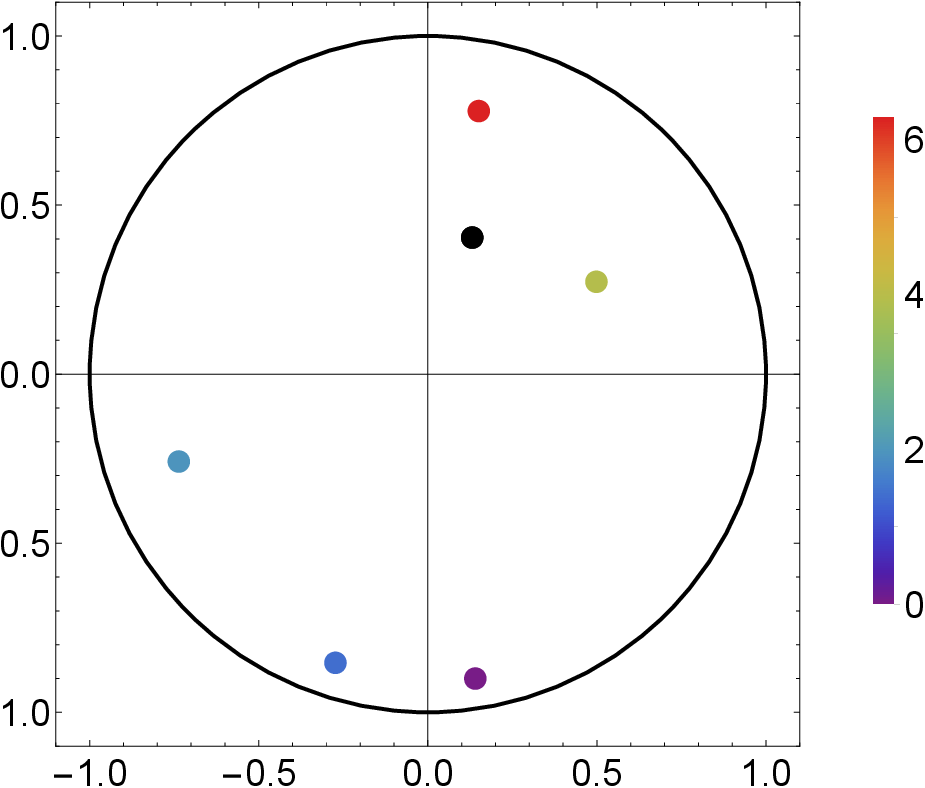}
  \end{tabular}
  \caption{\label{fig:2}Weighted barycenters of configurations of points in the Poincar\' e disc (two panels)}
\end{figure*}

Weighted barycenters for several configurations are plotted in Figure \ref{fig:2}.

The gradient descent algorithm in hyperbolic metric in $\mathbb{B}^n$ for computation of the weighted barycenter can be realized as follows.

\begin{enumerate}
\item[i)] Let $y_1,\dots,y_N$ be weighted observations with weights $w_1,\dots,w_N$ in $\mathbb{B}^n$. Suppose that $N>n$.

\item[ii)] Solve the system (\ref{swarm}) with $f$ defined as
$$
f = \frac{K}{N}\sum\limits_{j=1}^{N}\omega_jx_j, \quad K<0,
$$
and initial conditions $x_1(0)=y_1,\dots,x_N(0)=y_N$.

\item[iii)] For a sufficiently large $T$, it holds that $w_1 x_1(T)+\cdots + w_N x_N(T) = 0$. Compute the M\" obius transformation $g \in \mathbb{G}_n$, such that $x_1(T) = g(y_1),\dots,x_N(T) = g(y_N)$.

\item[iv)] The point $a_w=g^{-1}(0)$ is the weighted barycenter of points $y_1,\dots,y_N$.
\end{enumerate}

\begin{remark}
The system \eqref{swarm} in the Poincar\' e disc can be written as the system of complex-valued ODE's
\begin{equation}
\label{swarm_disc}
\dot z_i = i(f z_i^2 + \bar f), \; i=1,\dots,N \mbox{  where } f = \frac{iK}{2N} \sum_{i=1}^N z_i \mbox{  and } K<0.
\end{equation}
The notion $\bar f$ in the above expression stands for the conjugate of the complex number.
\end{remark}

\section{The M\" obius family of probability distributions in Poincar\' e balls}\label{sec:4}

Consider the following family of probability density functions in $\mathbb{B}^n$
\begin{equation}
\label{conf_nat_Poin_ball1}
p(x;a,s) = \frac{\pi^{n/2} \Gamma(1+s-n/2)}{\Gamma(1+s-n)} \left(\frac{(1-|x|^2)(1-|a|^2)}{\rho(a,x)}\right)^s, \quad x \in \mathbb{B}^d
\end{equation}
depending on parameters $a \in \mathbb{B}^n$ and $s>n-1$.

Following \cite{Jacimovic} we denote the family of probability distributions defined by densities \eqref{conf_nat_Poin_ball1} by $Moeb_n(a,s)$ and refer to them as {\it M\" obius distributions in the Poincar\' e ball}.

The family $Moeb_n(a,s^*)$ is conformally invariant for fixed $s^*$. This means that if $\mu \in Moeb_n(a,s^*)$ then $h_* \mu \in Moeb_n(a,s^*)$ for any $h \in \mathbb{G}_n$.

Moreover, the group $\mathbb{G}_n$ acts transitively on $Moeb_n(a,s^*)$. This means that for any $\mu_1,\mu_2 \in Moeb_n(a,s^*)$ there exists a transformation $h \in \mathbb{G}_n$, such that $h_* \mu_1 = \mu_2$.
\begin{figure*}[t]
\centering
  \begin{tabular}{@{}ccc@{}}
    \includegraphics[width=.3\textwidth]{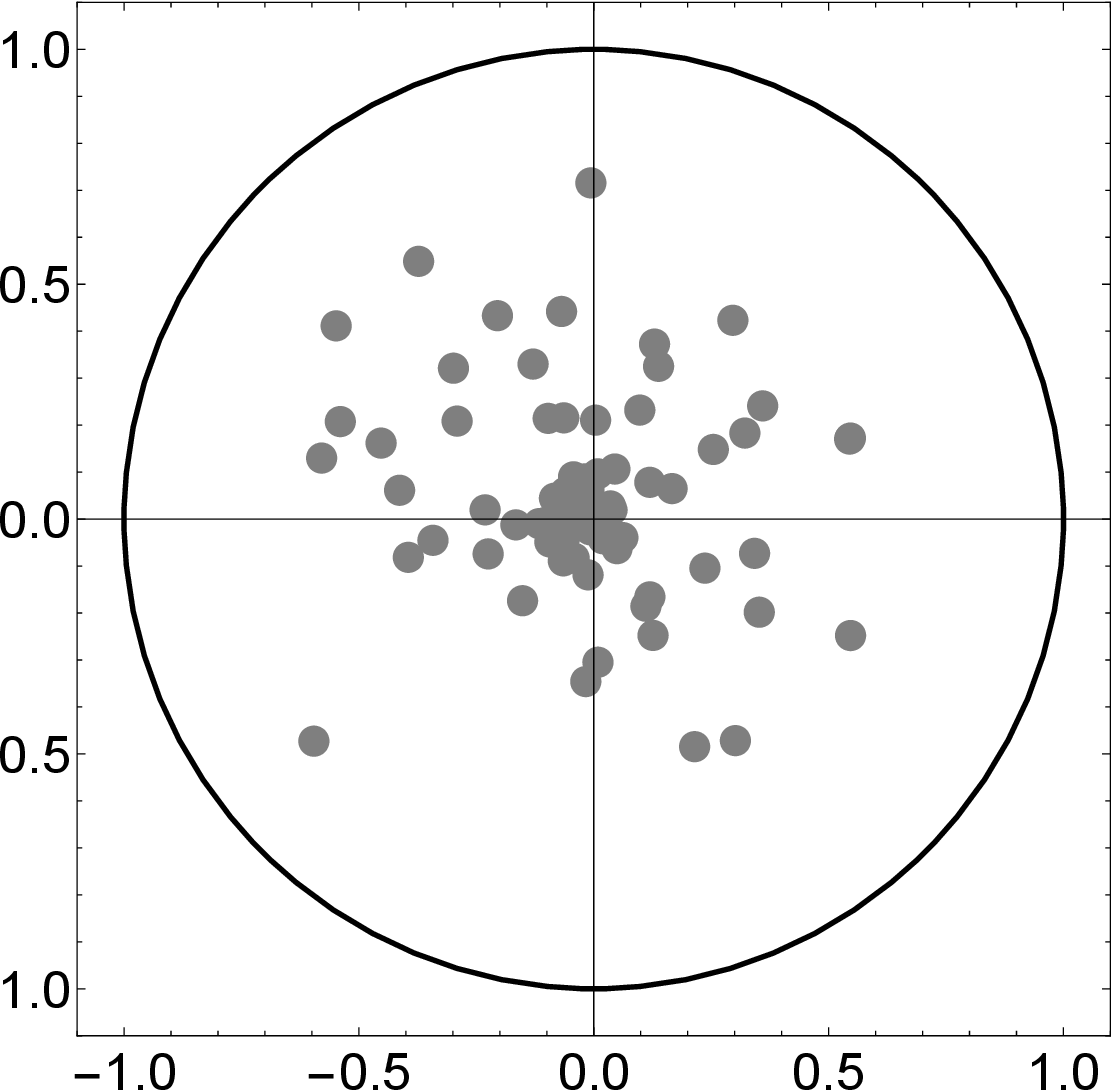}&\includegraphics[width=.3\textwidth]{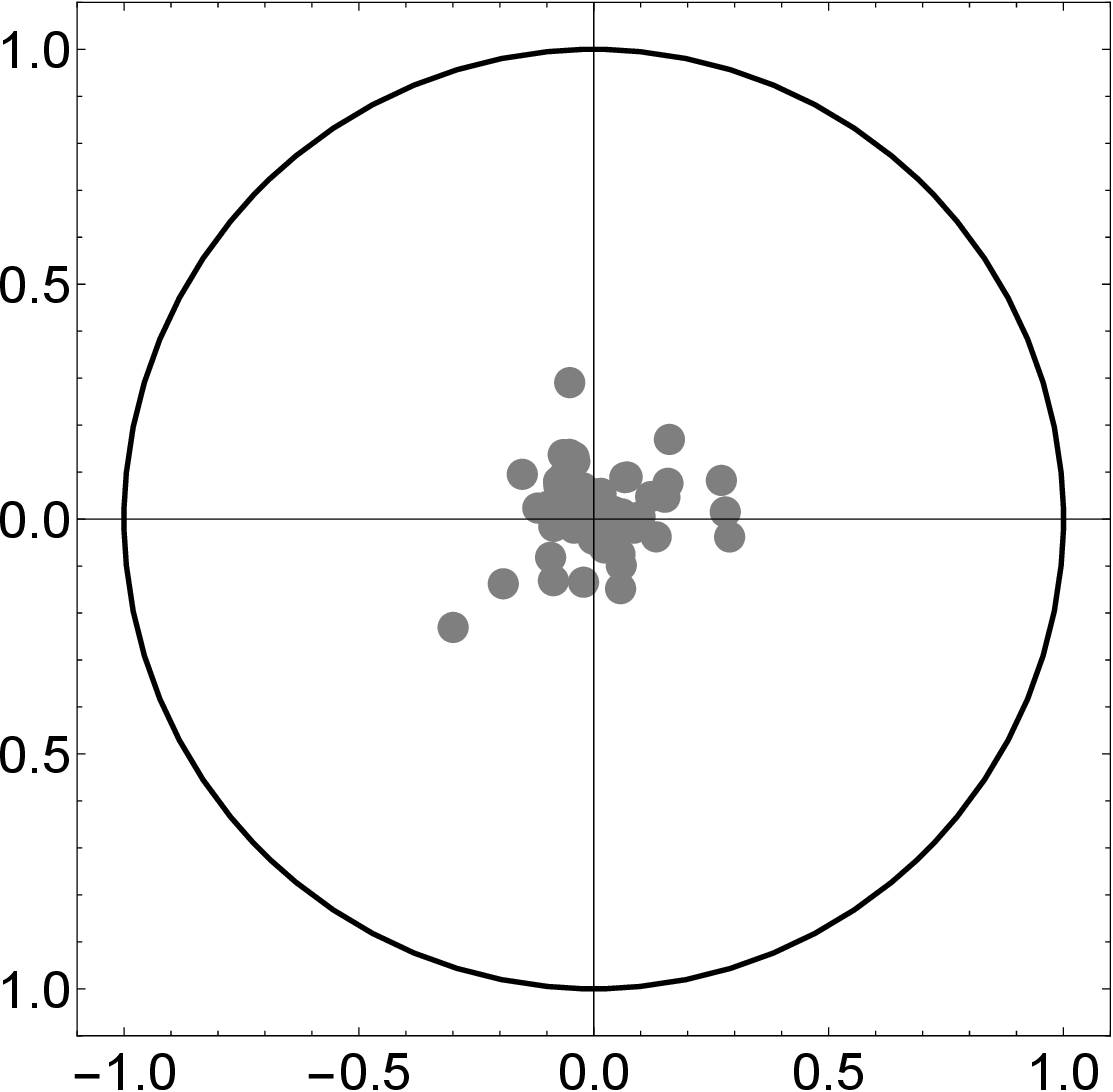}&\includegraphics[width=.3\textwidth]{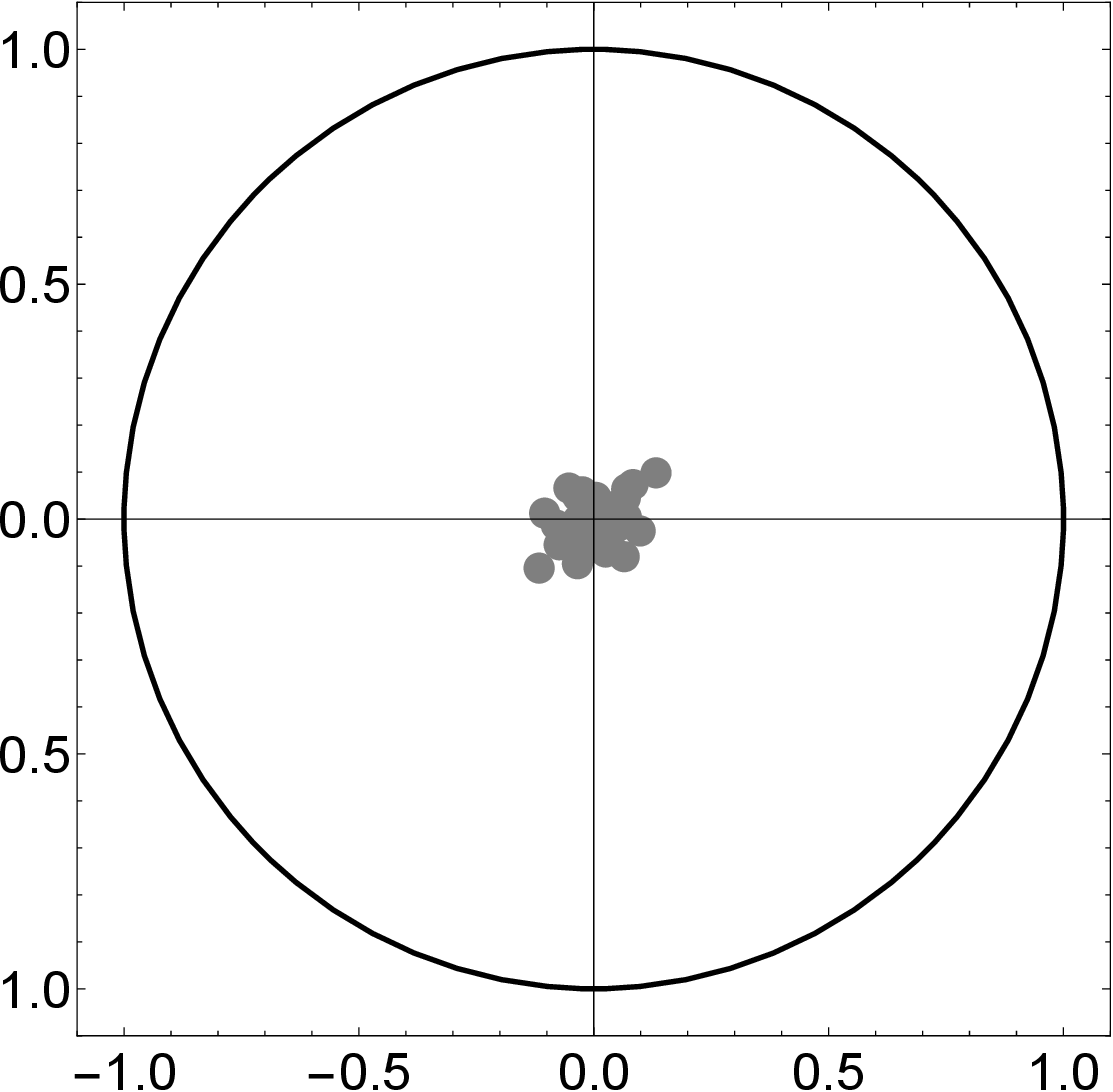}\\
    a) & b)  &c) \\
     \includegraphics[width=.3\textwidth]{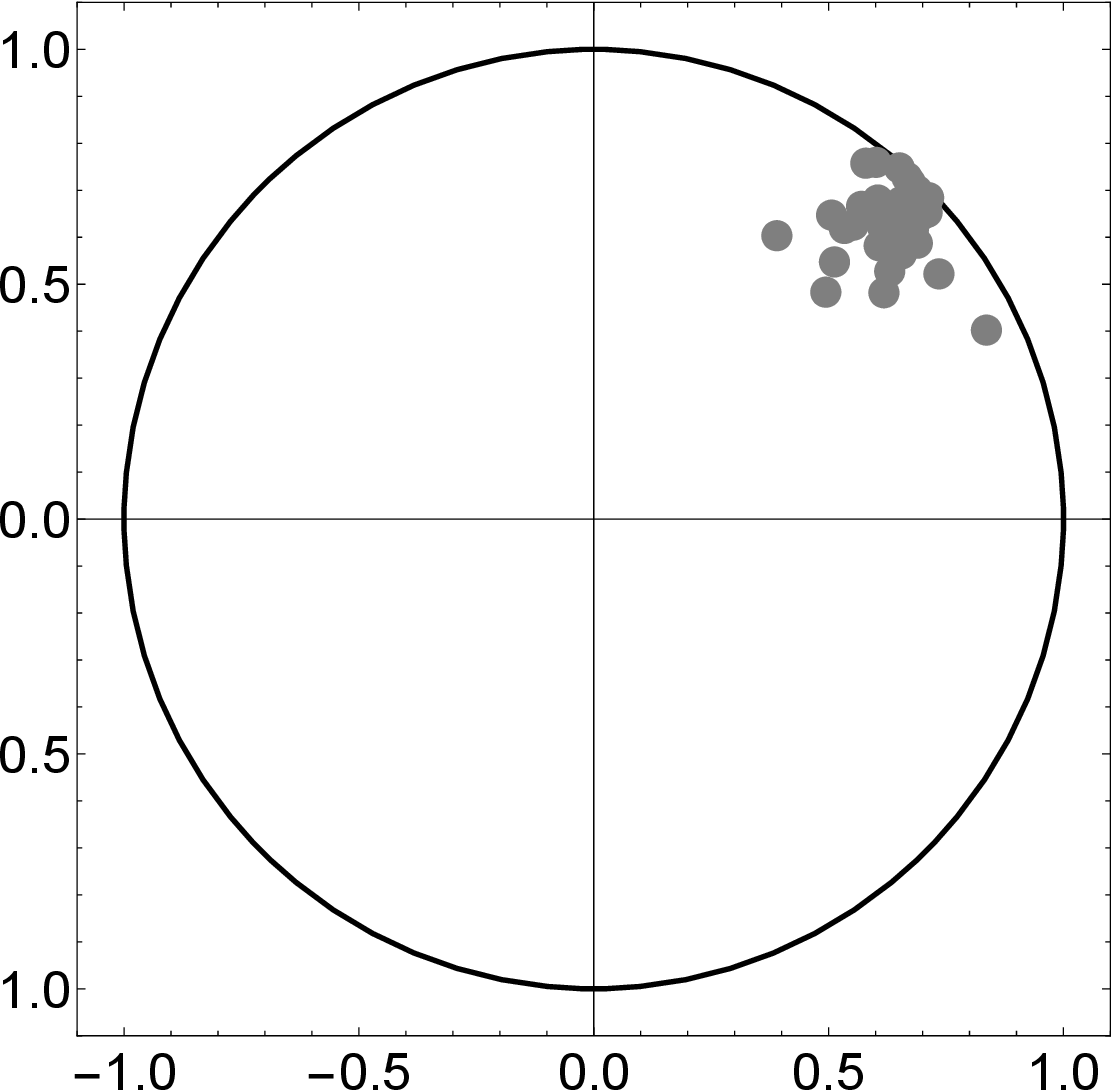}&\includegraphics[width=.3\textwidth]{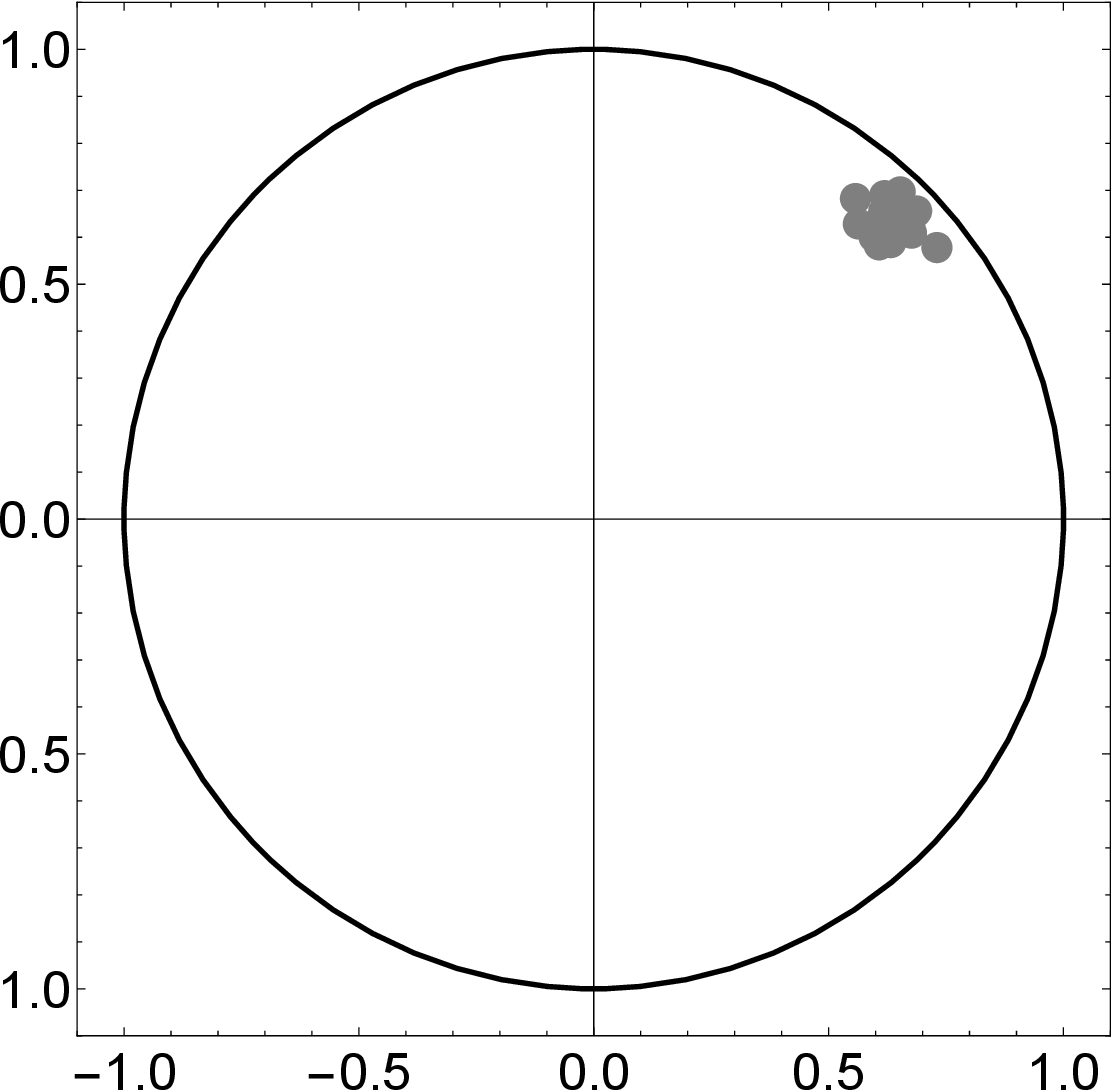}&\includegraphics[width=.3\textwidth]{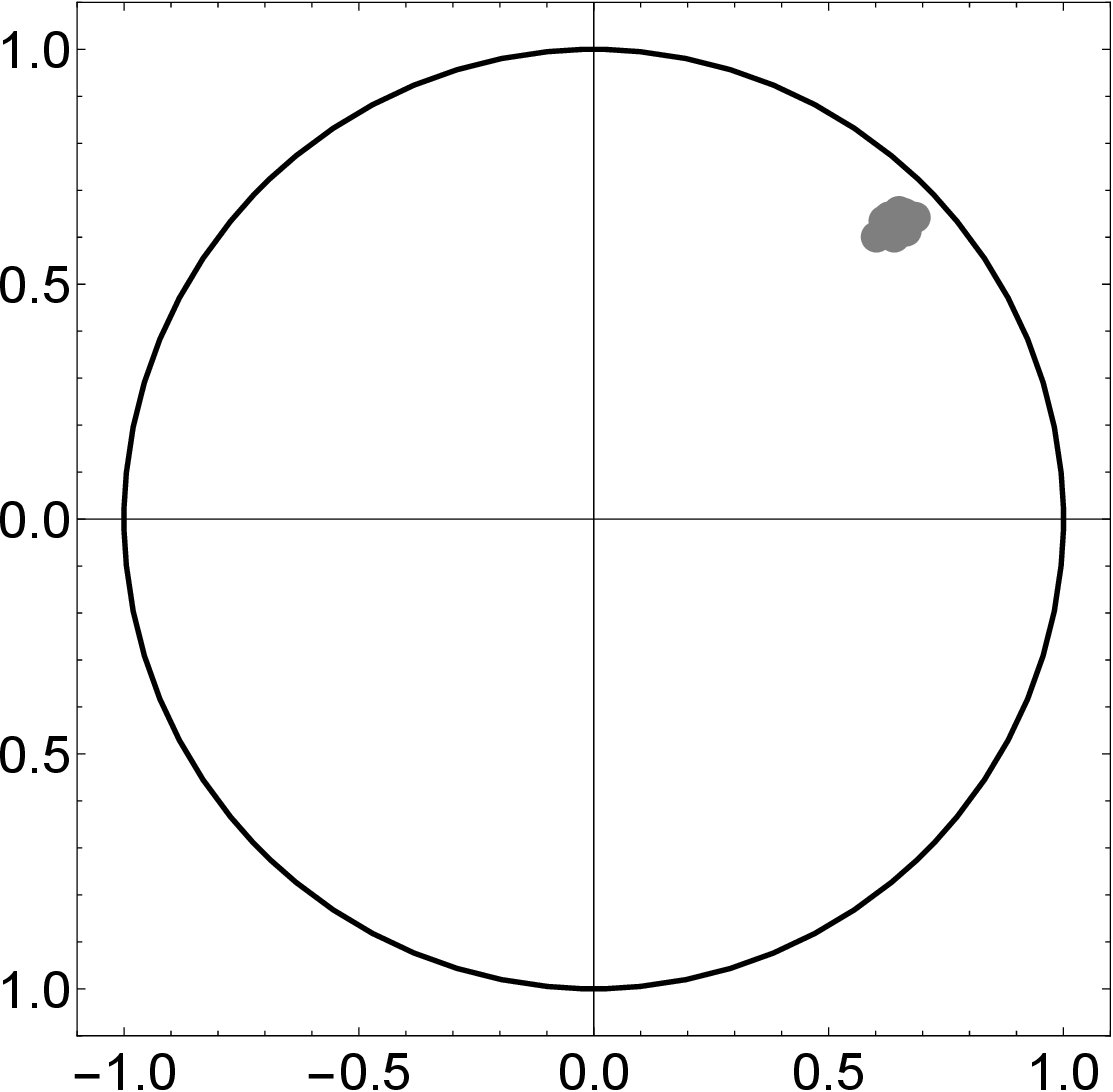}\\
    d) & e)  &f) \\
  \end{tabular}
  \caption{\label{fig:3}Random samples in Poincar\' e disc: a) $s=3$, $a=0$; b) $s=5$, $a=0$; c) $s=7$, $a=0$; d) $s=3$, $a=0.9e^{i\frac{\pi}{4}}$; e) $s=5$, $a=0.9e^{i\frac{\pi}{4}}$, and f) $s=7$, $a=0.9e^{i\frac{\pi}{4}}$.}
\end{figure*}

\subsection{Random variate generation}
\label{sampling}

For the sake of completeness we briefly expose the procedure for the random variate generation from the M\" obius distributions proposed in \cite{Jacimovic}. Suppose that we want to sample a random point from $Moeb(a^*,s^*)$, with given parameters $a^*$ and $s^*$.

\begin{enumerate}

\item[i)] Sample a uniformly distributed vector $u$ from the sphere ${\mathbb S}^{n-1}$ and a random number $\kappa$ uniformly distributed on $[0,1]$.

\item[ii)] Denote by $b^*$ the solution of the following equation:
\begin{equation}
\label{random_generate}
\frac{2 \Gamma(1+s^*-n/2)}{\Gamma(1+s^*-n)\Gamma(n/2)} \frac{b^n}{n} \; _2F_1(\frac{n}{2},n-s^*;\frac{n}{2}+1;b^2) = \kappa.
\end{equation}
The notation $_2 F_1(\cdot,\cdot;\cdot;\cdot)$ in the above equation stands for the Gaussian hypergeometric series.

\item[iii)] Then $y=b^* u$ is a random point in $\mathbb{B}^n$ distributed as $Moeb(0,s^*)$.

\item[iv)] Let $h \in \mathbb{G}_n$ be a M\" obius transformation, such that $h(0) = a^*$. Then $h(y) \sim Moeb(a^*,s^*)$.

\end{enumerate}

In Figure \ref{fig:3} we depict random samples from the M\" obius distributions in the Poincar\' e disc for different values of parameters $a$ and $s$.

\subsection{Maximum likelihood estimation}

 Given observations $y_1,\dots,y_N$ in $\mathbb{B}^n$, the maximum likelihood estimations $\hat a$ and $\hat s$ are (see \cite{Jacimovic}):
\begin{enumerate}
\item[a)] $\hat a$ is the conformal barycenter of points $y_1,\dots,y_N$;

\item[b)] Let $\hat a$ be MLE for the parameter $a$. Then $\hat s$ is the solution to the following optimization problem:

\begin{equation}
\label{s_MLE}
\mbox{ Maximize  } J(s) = \log \Gamma(s+1-\frac{n}{2}) - \log \Gamma(s+1-n) - \frac{s}{N} H_n(\hat a), \mbox{ w. r. to } s>n-1.
\end{equation}

It is easy to check that the above function is concave for $s>n-1$, hence, there is a unique solution of \eqref{s_MLE}.
\end{enumerate}

For the EM algorithm we need MLE for weighted observations from the family $Moeb(a,s)$. Let $y_1,\dots,y_N$ be observations in $\mathbb{B}^N$ with weights $w_1,\dots,w_N$. Then estimations for parameters $\hat a$ and $\hat s$ are given by
\begin{enumerate}
\item[a)] $\hat a$ is the weighted barycenter of points $y_1,\dots,y_N$.

\item[b)] $\hat s$ is solution to the following maximization problem:

\begin{equation}
\label{s_MLE_weights}
\mbox{ Maximize  } \tilde J(s) = \sum_{i=1}^N w_i \log \Gamma(s+1-\frac{n}{2}) - \sum_{i=1}^N w_i \log \Gamma(s+1-n) - \frac{s}{N} \tilde H_n(\hat a),
\end{equation}
where $\tilde H_n(\cdot)$ is defined by \eqref{weight_potential_Poin_ball}.
\end{enumerate}
\section{Clustering in the Poincar\' e disc} \label{sec:5}

As the necessary mathematical framework is introduced in sections \ref{sec:3} and \ref{sec:4}, everything is ready for implementation of the clustering algorithms.

We start with the Poincar\' e disc, as this two-dimensional manifold is convenient for visualization. In the next Section we extend the algorithms to Poincar\' e balls of arbitrary dimensions.

\subsection{k-means in the Poincar\' e disc}
\label{k-means_disc}

Let $\xi_1,\dots,\xi_N$ be observations in the Poincar\' e disc. Throughout this Section we use the algebra of complex numbers and complex-analytic tools. Therefore, we suppose that observations $\xi_i$ are complex numbers, such that $|\xi_i|<1$ for $i=1,\dots,N$. The notion $\bar \xi$ stands for the complex conjugate of $\xi$.

Suppose that there are $k$ clusters. Denote by $a_1,a_2,\ldots,a_k$ conformal barycenters of the clusters. The objective function of $k$-means clustering in the Poincar\' e disc is given by
$$J\left(\{a_j\},\{\kappa_{i,j}\}\right)=\sum\limits_{i=1}^{N}\sum\limits_{j=1}^{k}\kappa_{ij}d_{hyp}^{2}(\xi_i,a_j).$$
Here $d_{hyp}(\cdot,\cdot)$ is the hyperbolic distance in $\mathbb{B}^2$ and the binary variable $\kappa_{ij}\in\{0,1\}$ indicates whether the point $\xi_i$ belongs to the $j$-th cluster.

The $k$-means clustering in the Poincar\' e disc is implemented in a similar way as in the classical Euclidean case, with the two key differences: a) the use of hyperbolic distance; b) the use of conformal barycenters computed by the hyperbolic gradient descent algorithm.

\begin{enumerate}

\item[i)] Set parameters $a^* \in \mathbb{B}^2$ and $s^*>1$. In the general case (if no a priori information is available), $a^*=0$ and $s^* = 2$ are suggested values.

\item[ii)] Sample initial barycenters $a_1^{(0)},\dots,a_k^{(0)}$ in $\mathbb{B}^2$ from the probability distribution $Moeb_2(a^*,s^*)$.

\item[iii)] Assign each point $\xi_i$ to the nearest cluster barycenter:
$$\kappa_{ij}=\left\{\begin{array}{ll}
1& \textrm{if } d_{hyp}^{2}(\xi_i,a_j)=\min\limits_{1\leq j\leq k}d_{hyp}^{2}(\xi_i,a_j)\\
0&\textrm{otherwise.}
\end{array}\right.$$

\item[iv)] Find barycenters of all points assigned to each cluster by solving \eqref{swarm_disc} with the initial conditions $z_1(0) = \xi_1,\dots,z_N(0)=\xi_N$.

\item[v)] Repeat steps iii) and iv) until convergence.

\end{enumerate}

Experimental results are presented at the end of this Section, see figures \ref{fig:4}, \ref{fig:6}, \ref{fig:8} for three different sets of observations.

\subsection{Expectation-maximization for M\" obius mixture model in the Poincar\' e disc}

We now introduce expectation-maximization (EM) algorithm for the M\" obius mixture model in the Poincar\' e disc.

Let $\xi_1,\xi_2,\ldots,\xi_N$ be observations in the Poincar\' e disc.

\begin{enumerate}

\item[i)] {\bf Initialization}

 Take initial estimates for mixing probabilities $\pi_1^{(0)},\dots,\pi_k^{(0)}$, weighted hyperbolic barycenters $a_1^{(0)},\dots,a_k^{(0)}$ and concentration parameters $s_1^{(0)},\dots,s_k^{(0)}$.

In the absence of a prior information, one can set $\pi_i = 1/k$ for $i=1,\dots,k$; $a_i$ sampled from $Moeb_2(0,2)$ and $s_i = X+1$, where $X$ is the random number sampled from the exponential distribution with the expectation $1$.

\item[ii)] {\bf E-step}

ii-1) Compute the posterior probabilities for each point $\xi_i$
$$
\gamma_{im}=\frac{\pi_m p(\xi_i;a_m,s_m)}{\sum\limits_{j=1}^{k}\pi_j p(\xi_i;a_j,s_j)}, \quad m=1,\dots,k
$$
where $p(z;\cdot,\cdot)$ is the density function \eqref{conf_nat_Poin_ball1} for $n=2$:
$$
p(z;a,s)=\frac{s-1}{\pi}\left(\frac{(1-|a|^{2})(1-|z|^{2})}{|1-\bar a z|^{2}}\right)^{s},\; z \in\mathbb{B}^{2} \subset \mathbb{C}.
$$

ii-2) Compute weighted barycenters $a_m^{(j)}$ of points $\xi_i$, $i=1,\dots,N$ with weights $\gamma_{im}$ using the method explained in subsection \ref{weight_bary}.

Computation of each barycenter involves all the points in the data set, taken with their weights $\gamma_{im}$.

 \item[iii)] {\bf M-step}

iii-1) Update the estimates for mixing probabilities
$$
\pi_m^{(j)}=\frac{1}{N}\sum\limits_{i=1}^{N}\gamma_{im} \mbox{  for } m=1,\dots,k.
$$

iii-2) Weighted estimations for concentration parameters $s_m$ are solutions to the maximization problem \eqref{s_MLE} for the particular case $n=2$. This solution can be written in the closed form:
$$
s_m^{(j)} = 1 - \left( \sum \limits_{i=1}^N \gamma_{im}\right) \left(\sum\limits_{i=1}^N \gamma_{im}\;\textrm{log} \frac{(1-|a_m^{(j)}|^2)(1-|\xi_i|^2)}{|1-\bar \xi_i a_m^{(j)}|^2} \right)^{-1} \mbox{  for  } m=1,\dots,k.
$$

\item[iv)] {\bf Iteration}

Iterate steps ii) and iii) for $j=1,\dots$ until convergence.
\end{enumerate}

\subsection{Experiments}

In this subsection we present results of the $k$-means method and EM algorithm for three data sets in the Poincar\' e disc. The observations are generated from the M\" obius mixture model in the Poincar\' e disc. Hence, one can check if the EM algorithm recovers parameters of the mixture model from which the data are generated.

All the experiments have been implemented in Wolfram Mathematica 11.0.

\subsubsection{Experiment A1}

\noindent{\it Generation of the data set}

In this experiment the data are generated from the mixture of M\" obius distributions with the following parameters:
\begin{enumerate}
\item[a)] The first cluster: $\pi_1=3/10$, $a_1=0$, $s_1=5$;

\item[b)] The second cluster: $\pi_2 =3/10$, $a_2=3/4$, $s_2=2$;

\item[c)] The third cluster: $\pi_3=1/4$, $a_3=-2i/3$, $s_3=8$;

\item[d)] The forth cluster: $\pi_4=3/20$, $a_4=1/2+i/2$, $s_4=10$.
\end{enumerate}

\noindent{\it Results}

K-means found the following barycenters: $a_1=-0.0000426429-0.0070095 i$, $a_2=0.785931-0.0545432 i$, $a_3=0.00336615-0.666576 i$ and $a_4=0.524064+0.46399 i$.

EM algorithm found the following mixture of the M\" obius distributions in $\mathbb{B}^2$:
\begin{enumerate}
\item[a)] The first cluster: $\pi_1=0.301478$, $a_1=-0.010205-0.00648238 i$, $s_1=5.01062$;

\item[b)] The second cluster: $\pi_2 =0.288929$, $a_2=0.765946-0.0342974 i$, $s_2=1.865$;

\item[c)] The third cluster: $\pi_3=0.249908$, $a_3=0.00336701-0.66657 i$, $s_3=6.96783$;

\item[d)] The forth cluster: $\pi_4=0.159685$, $a_4=0.503879+0.500164 i$, $s_4=6.42473$.
\end{enumerate}

\begin{figure*}[t]
\centering
  \begin{tabular}{@{}ccc@{}}
    \includegraphics[width=.3\textwidth]{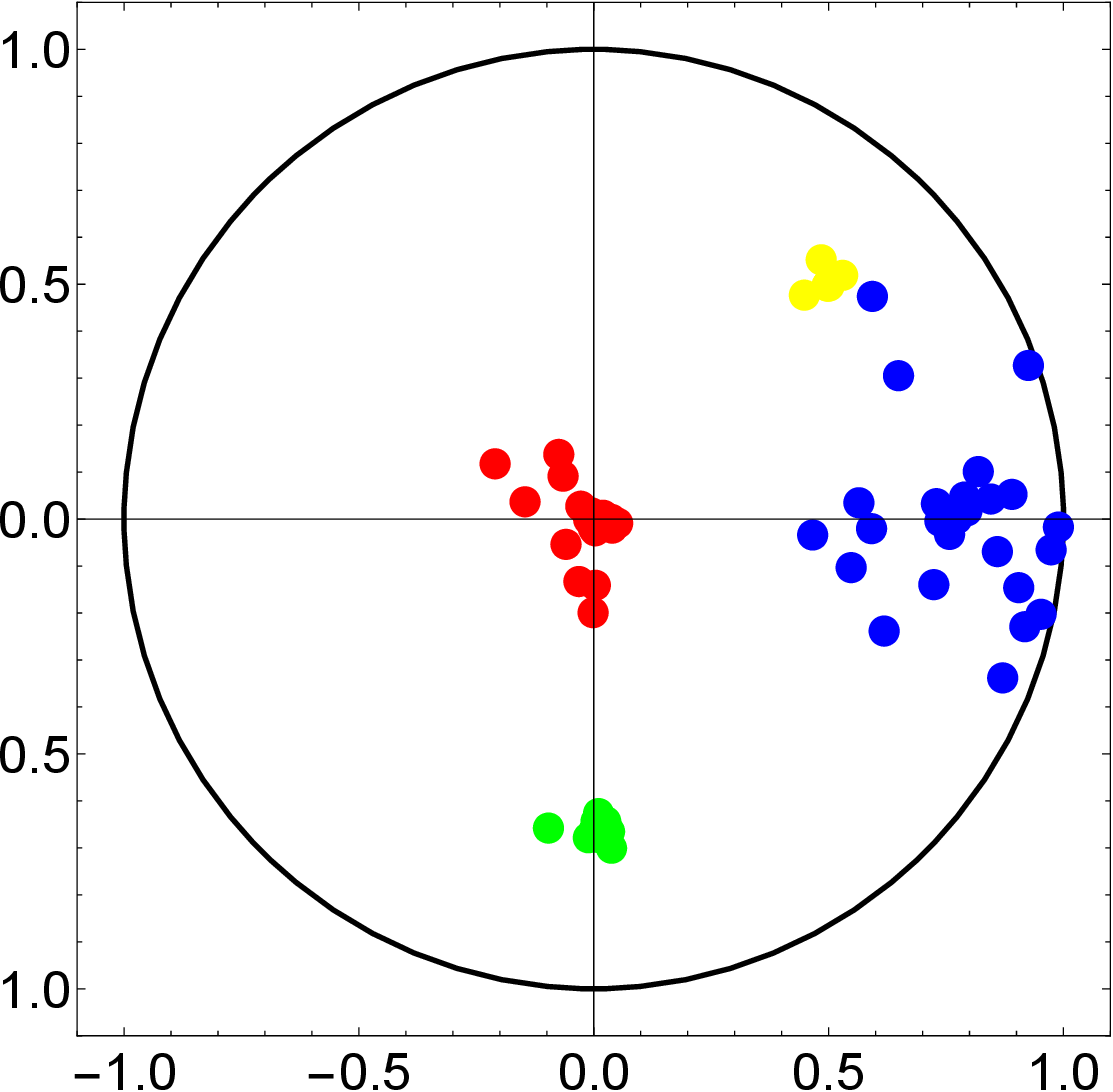}&\includegraphics[width=.3\textwidth]{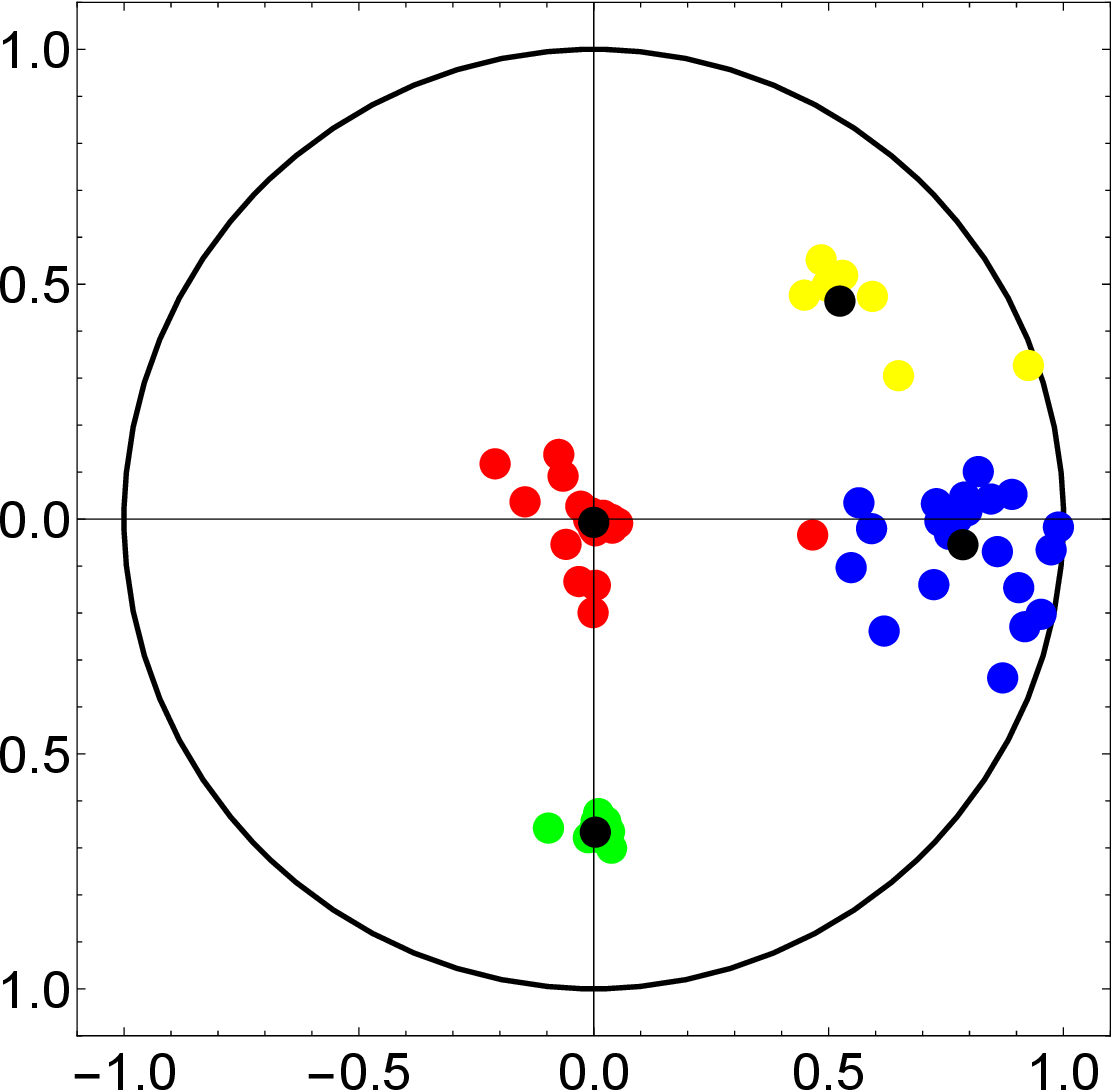}&\includegraphics[width=.3\textwidth]{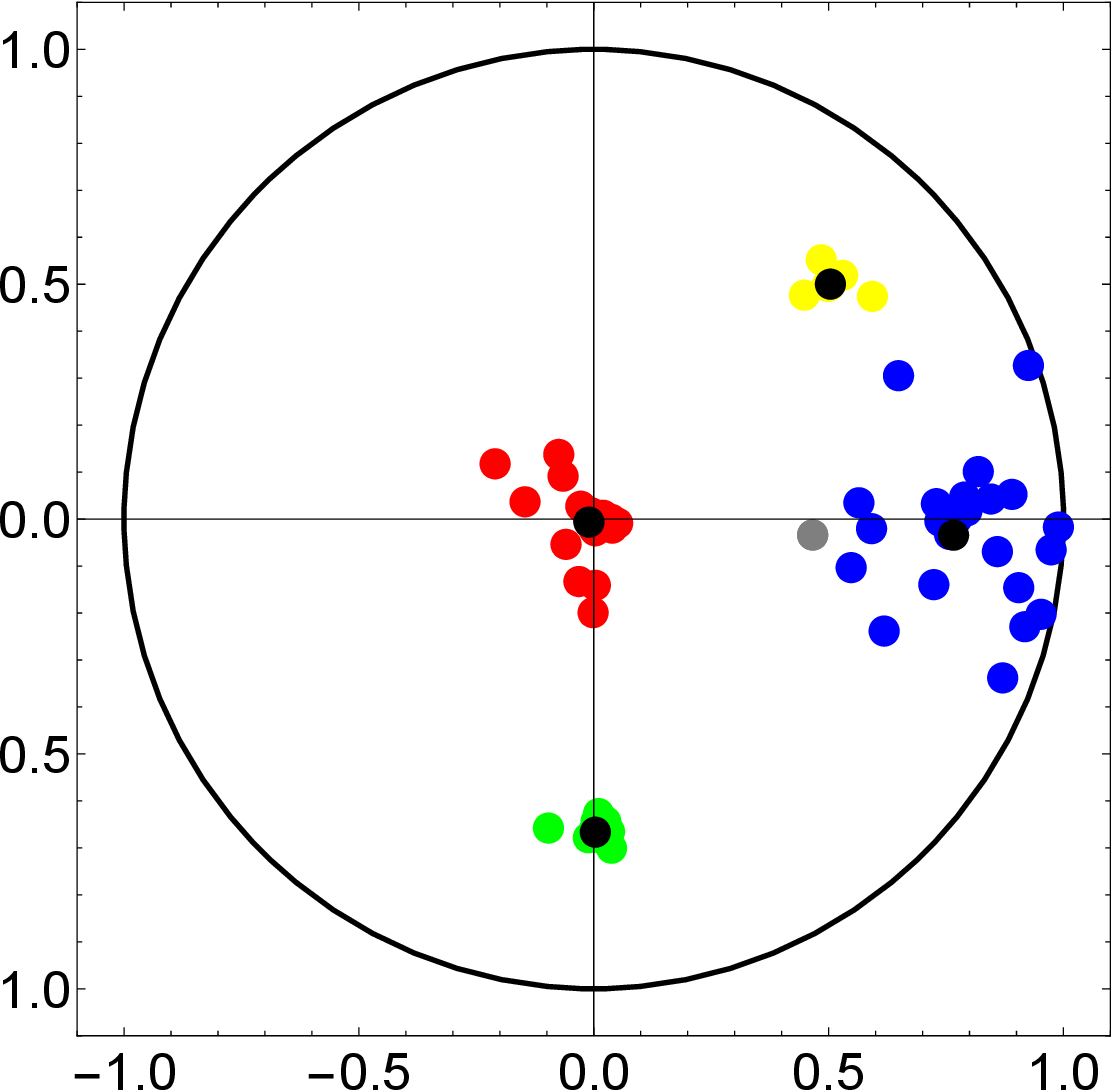}\\
    a) & b)  &c) \
  \end{tabular}
  \caption{\label{fig:4}Experiment A1 (three panels): a) ground truth; b) k-means; c) EM algorithm}
\end{figure*}

\begin{figure*}[t]
\centering
  \begin{tabular}{@{}c@{}}
    \includegraphics[width=.4\textwidth]{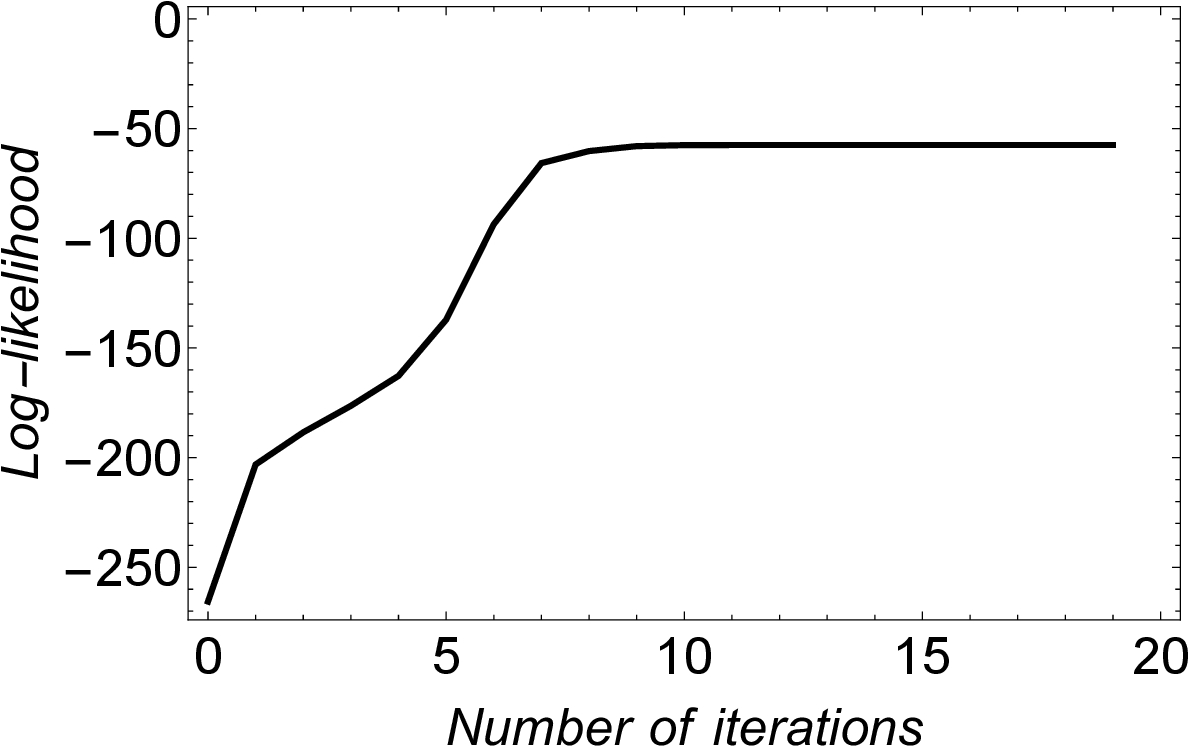}
  \end{tabular}
  \caption{\label{fig:5}Log-likelihood for EM in Experiment A1}
\end{figure*}

Results of Experiment A1 are visualized in Figure \ref{fig:4}. Here and below, the points that the EM algorithm did not decisively assigned to any cluster are shown in gray. As we can see there is one gray point in Figure \ref{fig:4}c).

Evolution of the log-likelihood function in the EM algorithm is plotted in Figure \ref{fig:5}.

\subsubsection{Experiment A2}

\noindent{\it Generation of the data set}

The observations are generated from the following mixture of M\" obius distributions on $\mathbb{B}^2$:
\begin{enumerate}
\item[a)] The first cluster: $\pi_1=1/5$, $a_1=0.95$, $s_1=7$;

\item[b)] The second cluster: $\pi_2 =1/4$, $a_2=0.9-0.15i$, $s_2=9$;

\item[c)] The third cluster: $\pi_3=3/20$, $a_3=0$, $s_3=4$;

\item[d)] The forth cluster: $\pi_4=2/5$, $a_4=-0.5$, $s_4=1.5$.
\end{enumerate}

\noindent{\it Results}

K-means found: $a_1=0.891161-0.0656324 i$, $a_2=-0.8407+0.139705 i$, $a_3=-0.209848+0.0666284 i$ and $a_4=-0.618712-0.456431 i$.

EM found the following mixture of M\" obius distributions:
\begin{enumerate}
\item[a)] The first cluster: $\pi_1=0.124334$, $a_1=-0.501575-0.0387122 i$, $s_1=3.17644$;

\item[b)] The second cluster: $\pi_2=0.474575$, $a_2=0.888033-0.0660443 i$, $s_2=1.80626$;

\item[c)] The third cluster: $\pi_3=0.139317$, $a_3=-0.0031058+0.0170453 i$, $s_3=4.3573$;

\item[d)] The forth cluster: $\pi_4 =0.261774$, $a_4=-0.512919-0.0327357 i$, $s_4=1.36019$.
\end{enumerate}

\begin{figure*}[t]
\centering
  \begin{tabular}{@{}ccc@{}}
    \includegraphics[width=.3\textwidth]{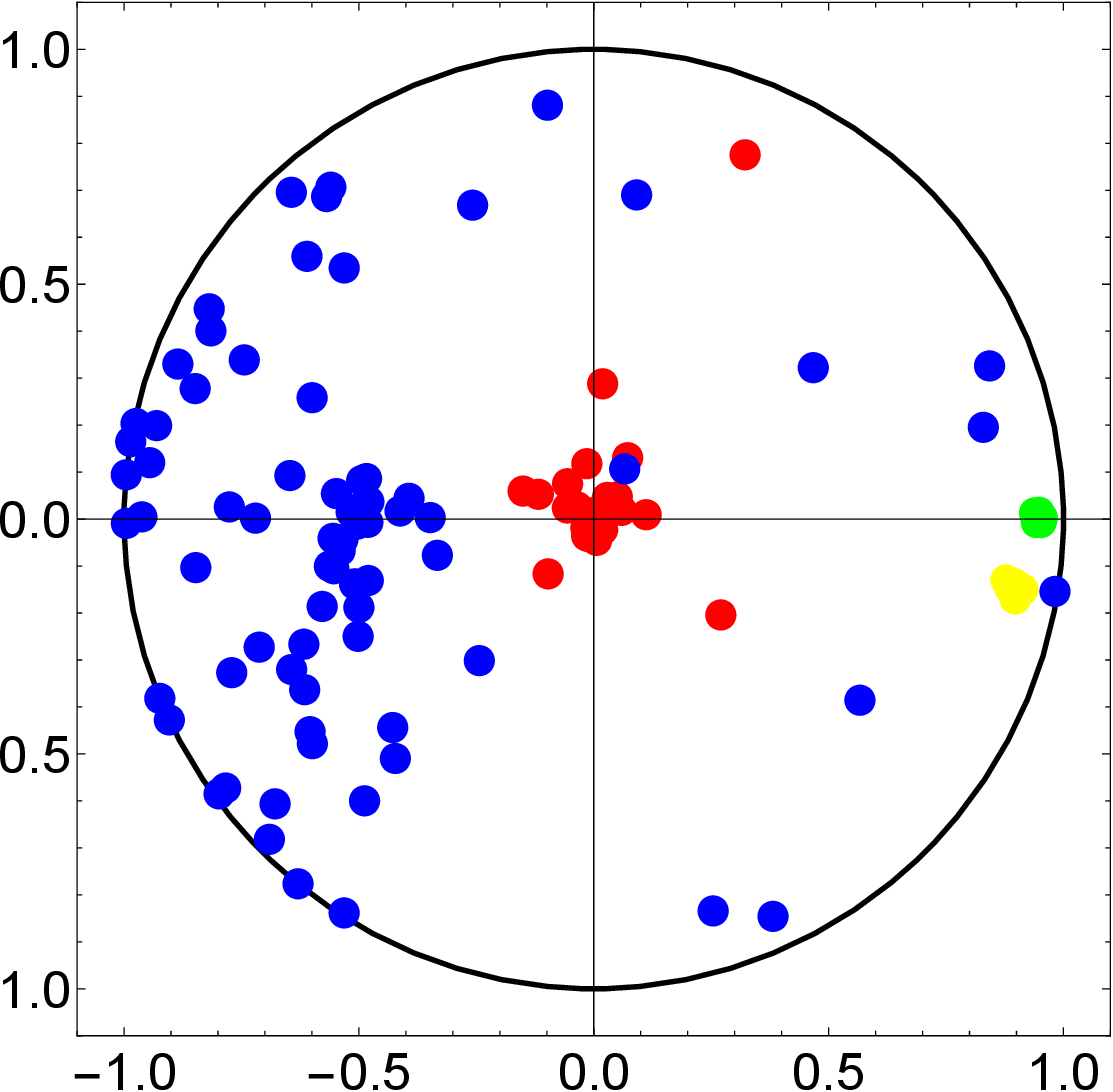}&\includegraphics[width=.3\textwidth]{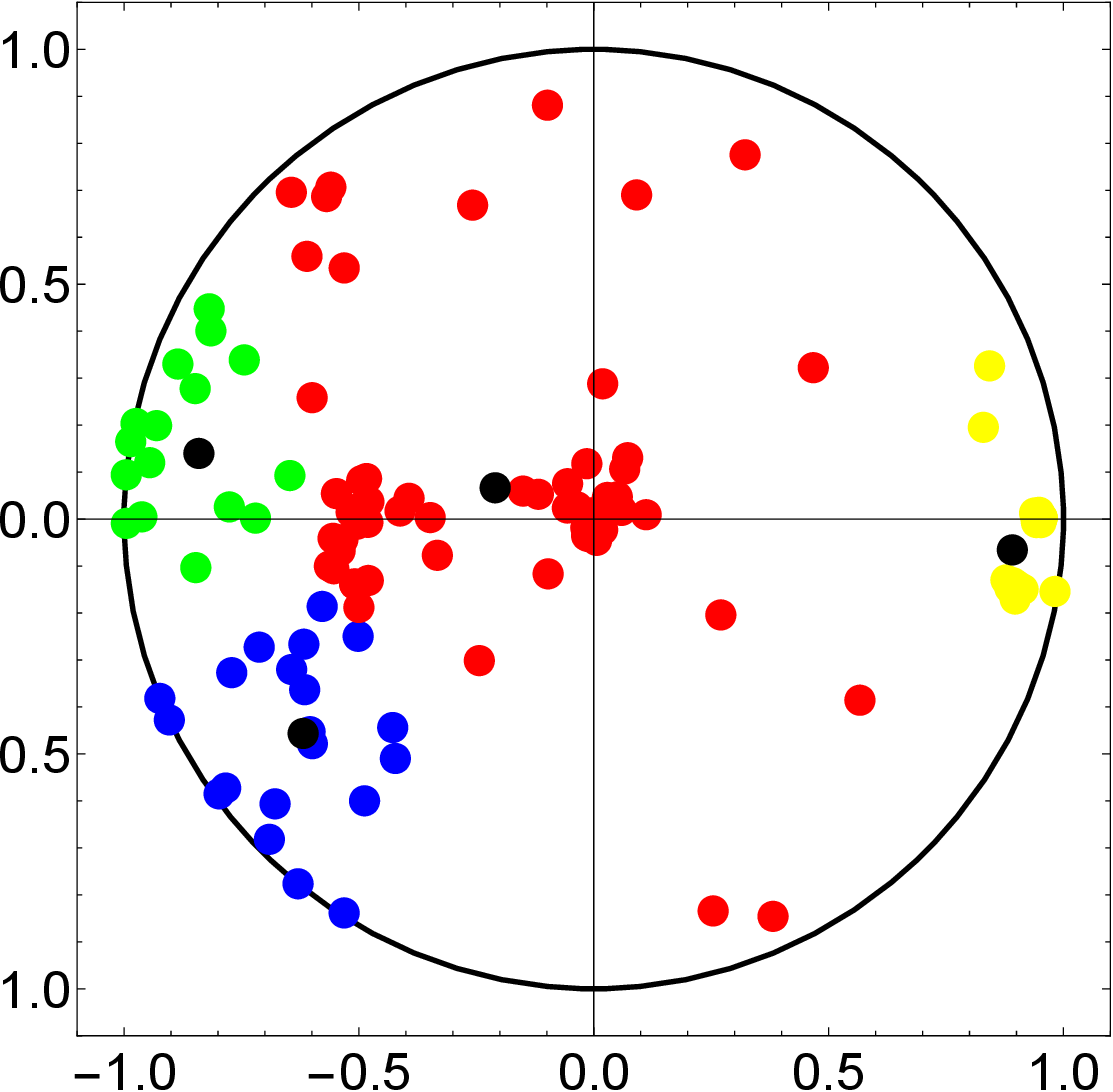}&\includegraphics[width=.3\textwidth]{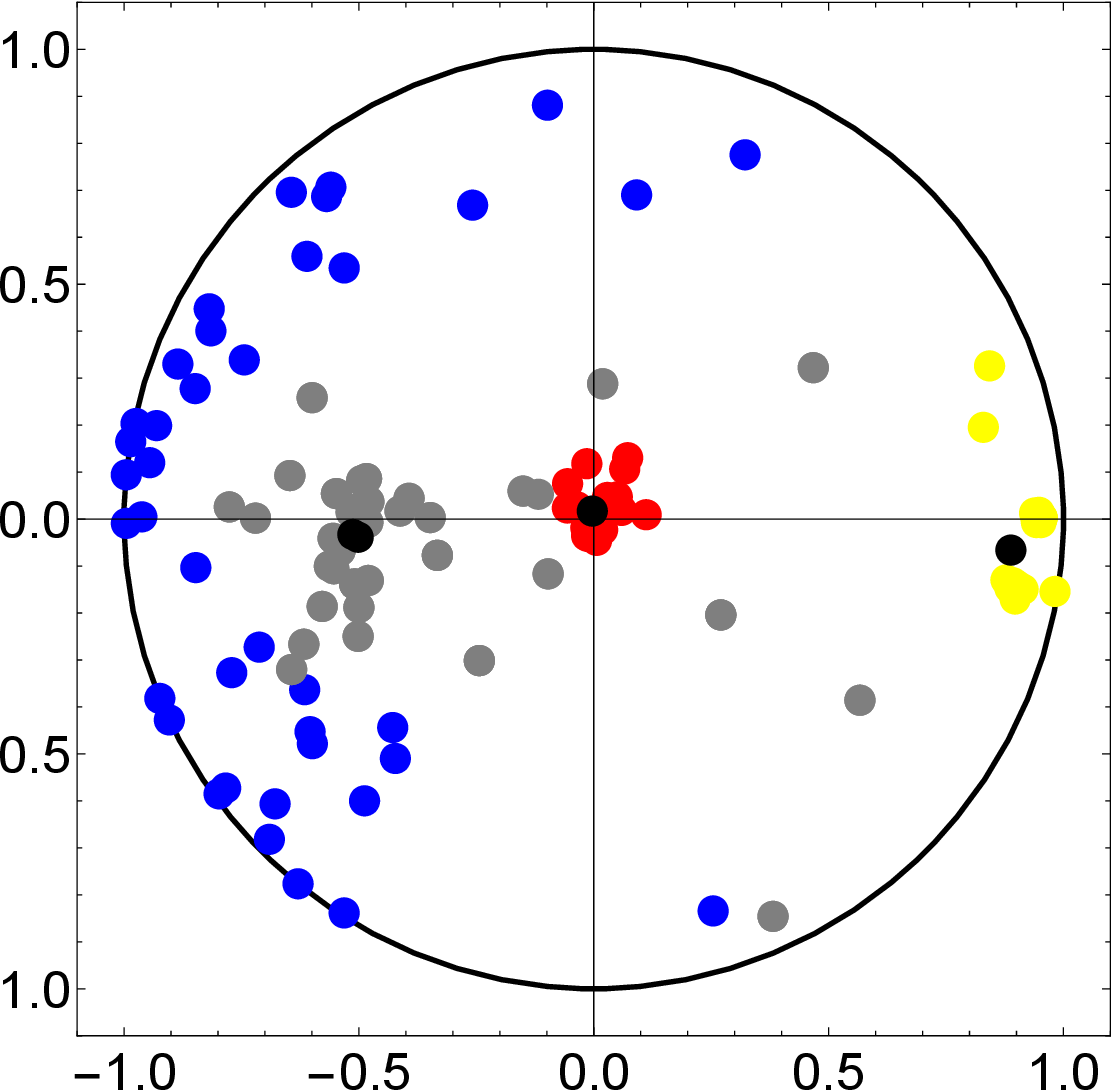}\\
    a) & b)  &c) \
  \end{tabular}
  \caption{\label{fig:6}Experiment A2 (three panels): a) ground truth; b) k-means; c) EM algorithm}
\end{figure*}

\begin{figure*}[t]
\centering
  \begin{tabular}{@{}c@{}}
    \includegraphics[width=.4\textwidth]{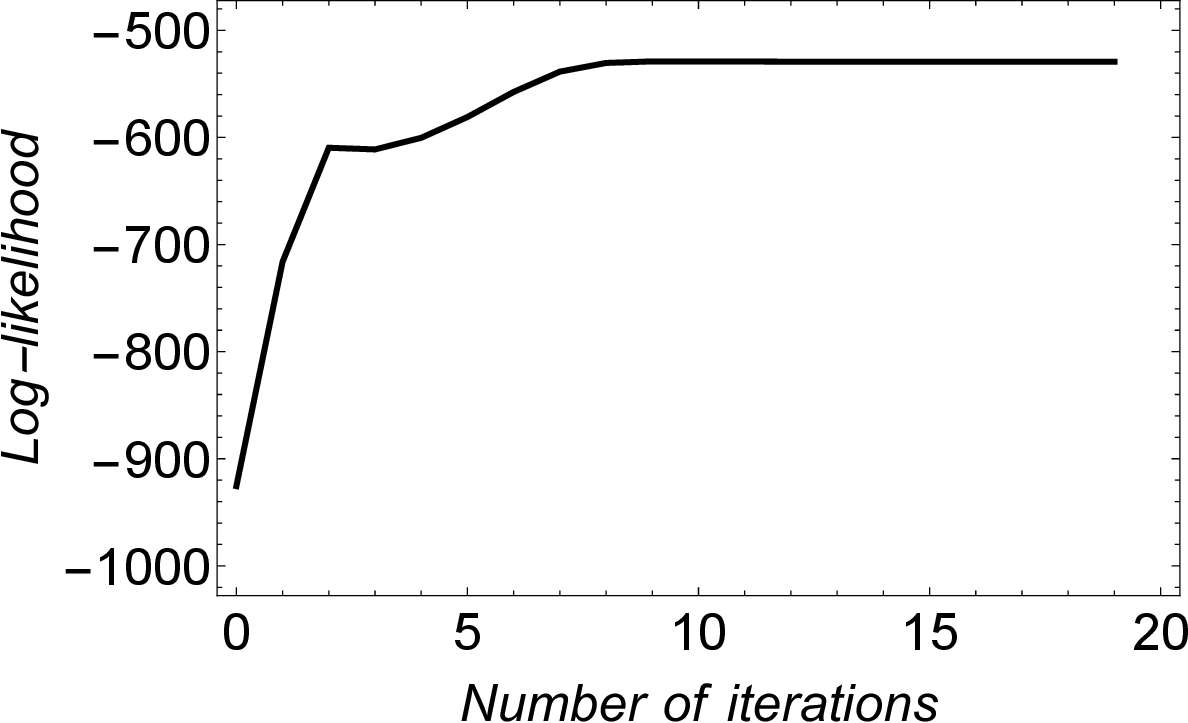}
  \end{tabular}
  \caption{\label{fig:7}Log-likelihood for EM in Experiment A2}
\end{figure*}

Results of Experiment A2 are visualized in Figure \ref{fig:6}. One can notice that both algorithms suggested the clusters which significantly differ from the ground truth. Moreover, EM algorithm essentially merged two barycenters and left many points not decisively assigned to a specific cluster (gray points in Figure \ref{fig:6}c).

Evolution of the log-likelihood function in the EM algorithm is shown in Figure \ref{fig:7}.

\subsubsection{Experiment A3}

\noindent{\it Generation of the data set}

Data are generated from the mixture of three M\" obius distributions:
\begin{enumerate}
\item[a)] The first cluster: $\pi_1=7/20$, $a_1=0$, $s_1=3$;

\item[b)] The second cluster: $\pi_2 =1/5$, $a_2=1/4$, $s_2=4$;

\item[c)] The third cluster: $\pi_3=9/20$, $a_3=1/3-i/3$, $s_3=5$.
\end{enumerate}

\noindent{\it Results}

K-means found the following barycenters: $a_1=-0.023886-0.0199269 i$, $a_2=0.247391+0.0562165 i$ and $a_3=0.338258-0.334991 i$.

EM did not recognize any clusters in this data sets. Instead, the algorithm suggested that all observations come from a single M\" obius distributions with parameters $a_{1,2,3}=0.20577-0.168234 i$ and $s_{1,2,3}=2.37012$.

\begin{figure*}[t]
\centering
  \begin{tabular}{@{}ccc@{}}
    \includegraphics[width=.3\textwidth]{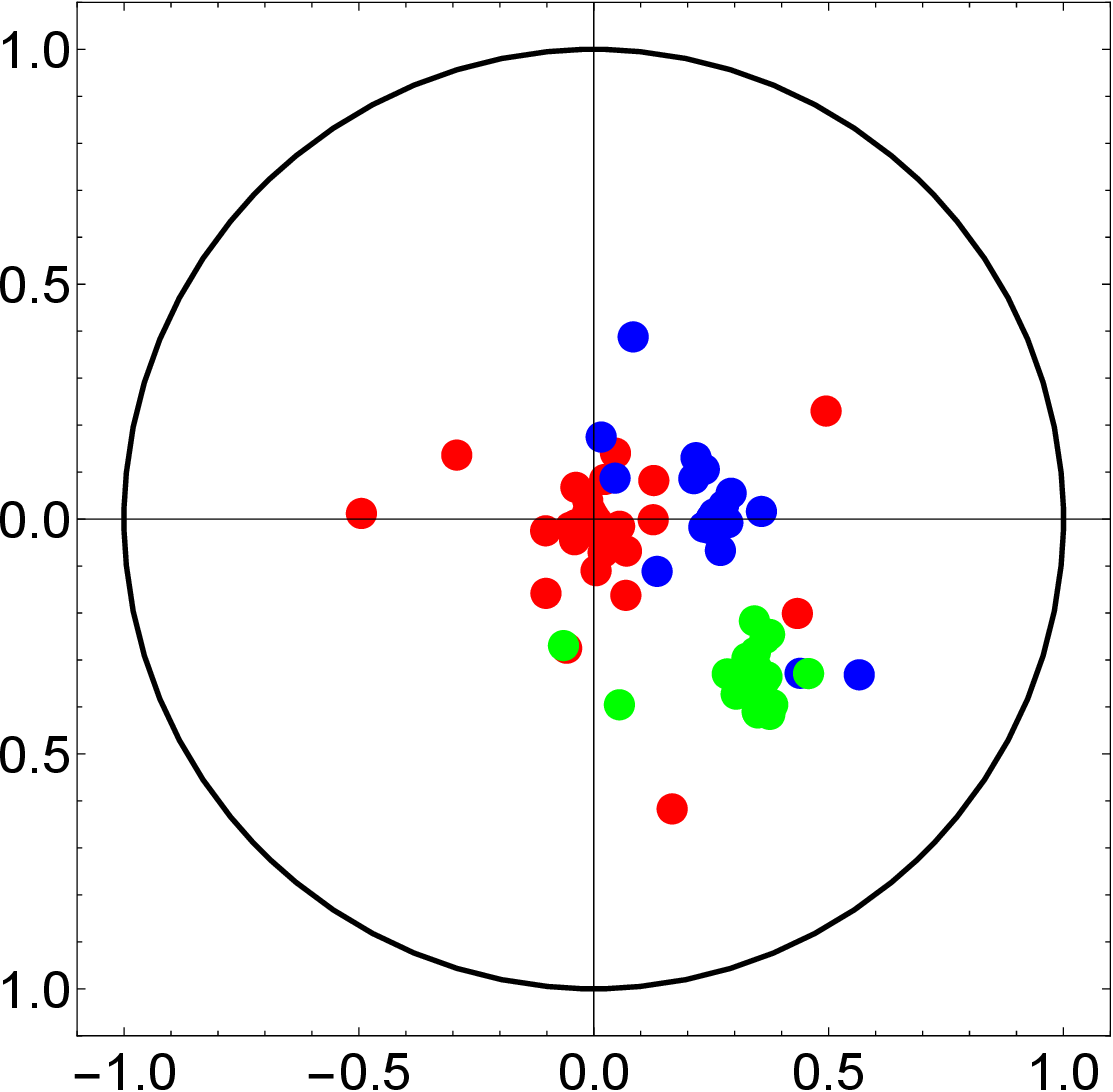}&\includegraphics[width=.3\textwidth]{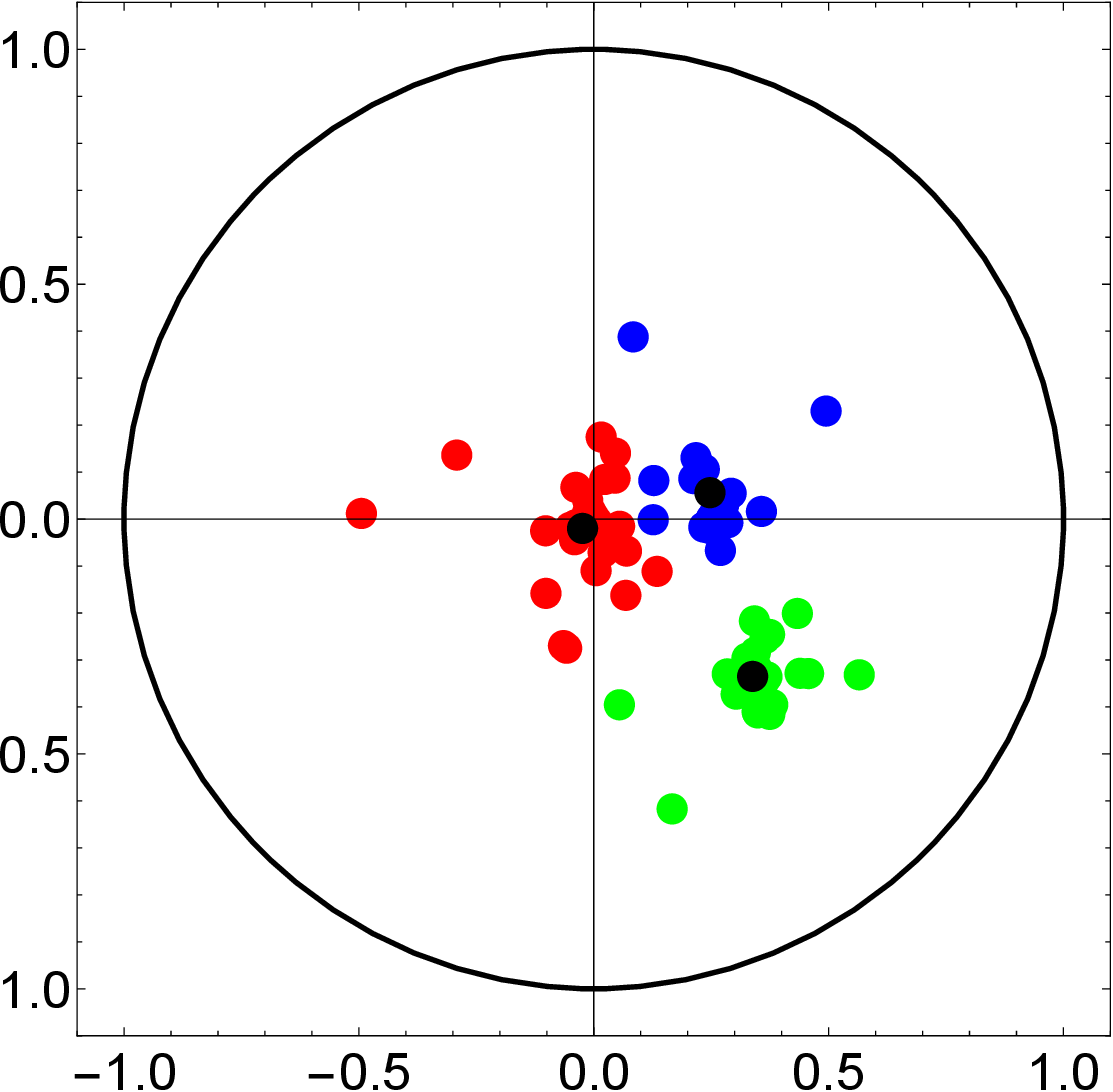}&\includegraphics[width=.3\textwidth]{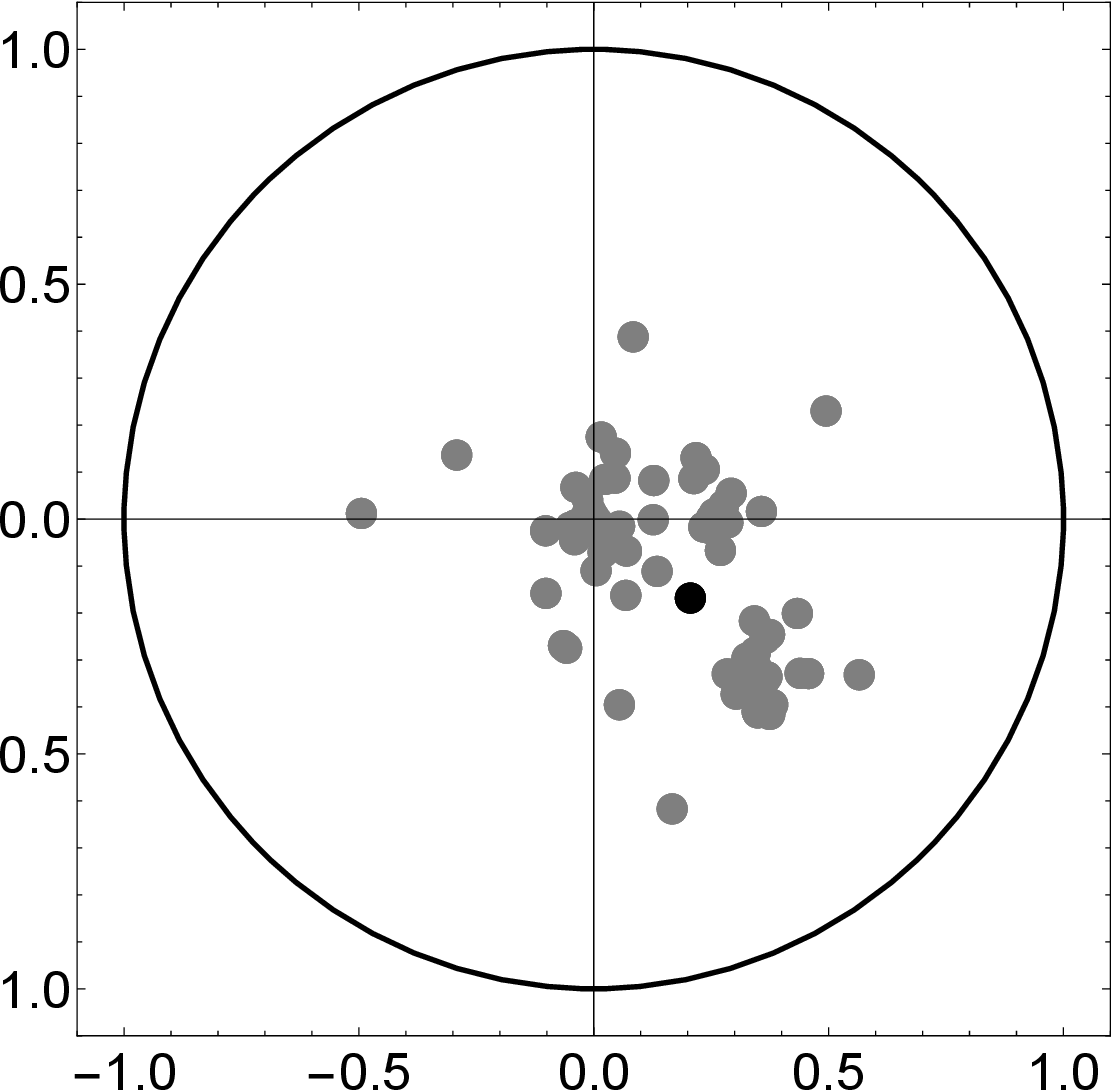}\\
    a) & b)  &c) \
  \end{tabular}
  \caption{\label{fig:8}Experiment A3 (three panels): a) ground truth; b) k-means; c) EM algorithm}
\end{figure*}

Results are visualized in Figure \ref{fig:8}. As we can see, all points in the right panel are colored in gray.

\begin{figure*}[t]
\centering
  \begin{tabular}{@{}c@{}}
    \includegraphics[width=.4\textwidth]{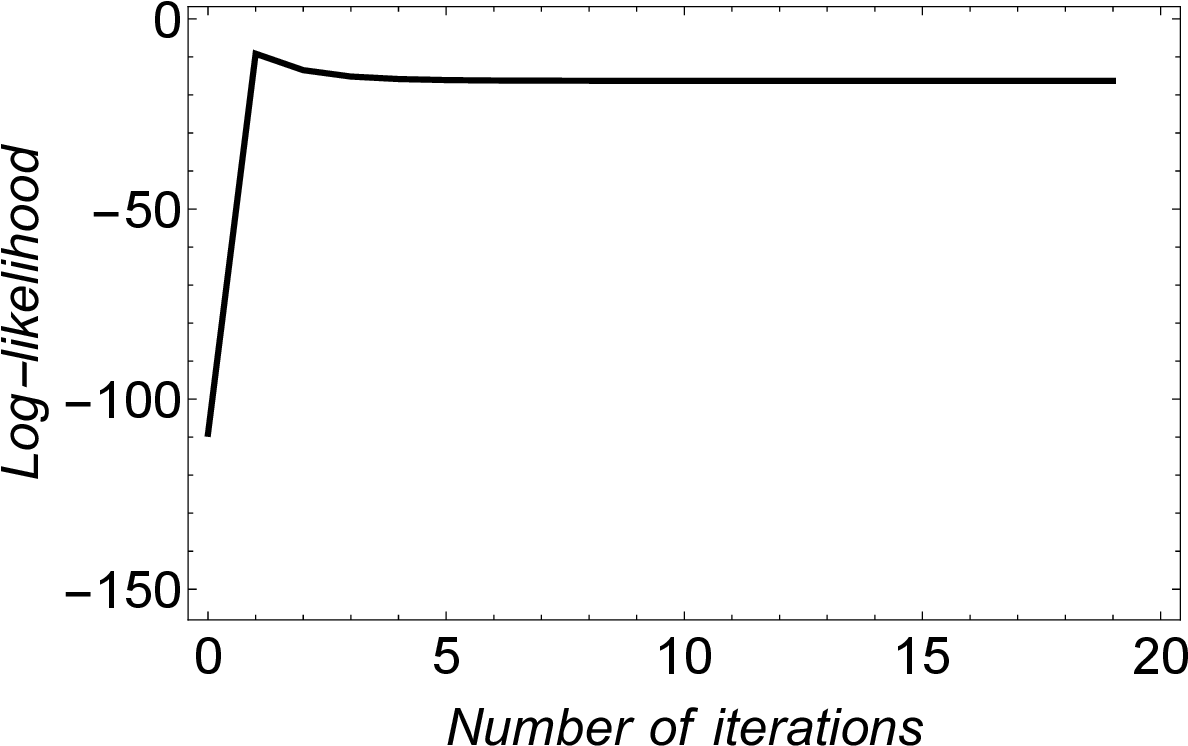}
  \end{tabular}
  \caption{\label{fig:9}Log-likelihood for EM in Experiment A3}
\end{figure*}

Evolution of the log-likelihood function in the EM algorithm is shown in Figure \ref{fig:9}.

\section{Clustering in Poincar\' e balls} \label{sec:6}

In the present Section we explain clustering algorithms for Poincar\' e balls in any dimension. We will also present experiments in the three-dimensional Poincar\' e ball $\mathbb{B}^3$.

\subsection{k-means in Poincar\' e balls}

K-Means for clustering in Poincar\' e balls is analogous to the algorithm in the Poincar\' e disc (explained in subsection \ref{k-means_disc}). The main difference is that we use the system \eqref{swarm} for computation of weighted conformal barycenter.

Experimental results in the three-dimensional case are presented below in figures \ref{fig:10} and \ref{fig:12}.

\subsection{EM for learning mixtures in Poincar\' e balls}

Let $y_1,\dots,y_N$ be observations in the Poincar\' e ball $\mathbb{B}^n$.
\newpage
\begin{enumerate}

\item[i)]  {\bf Initialization}

Take initial estimates for mixing probabilities $\pi_m^{(0)}$, weighted hyperbolic barycenters $a_m^{(0)}$ and concentration parameters $s_m^{(0)}$ for $m=1,\dots,k$.

\item[ii)] {\bf E-step}

ii-1) Compute the responsibilities for each data point $y_i$
$$
\gamma_{im}=\frac{\pi_m p(y_i;a_m,s_m)}{\sum\limits_{j=1}^{k}\pi_j p(y_i;a_j,s_j)}, \quad m=1,\dots,k
$$
where $p(x;\cdot,\cdot)$ is density function \eqref{conf_nat_Poin_ball1}.


ii-2) Compute $k$ weighted barycenters $a_m^{(j)}, \, m=1,\dots,k$ of points $y_i$, $i=1,\dots,N$ with weights $\gamma_{im}$ obtained in the previous step. The method for computation is exposed in subsection \ref{weight_bary}.

\item[iii)] {\bf M-step}

iii-1) Update the estimates for mixing probabilities
$$
\pi_m^{(j)}=\frac{1}{N}\sum\limits_{i=1}^{n}\gamma_{im}.
$$

iii-2) Estimations for concentration parameters $s_m^{(j)}$ are maximizers of the function \eqref{s_MLE_weights}.

\item[iv)] {\bf Iteration}

Iterate steps ii) and iii) for $j=1,\dots$ until convergence.
\end{enumerate}

\begin{remark}
At each iteration of the algorithm it is required to solve the maximization problem \eqref{s_MLE_weights} in the step iii-2). The objective function is concave and the optimization problem is easy for the numerical solution, especially due to the fact that $n$ is an integer. The logarithmic derivative of the Gamma function (also referred to as digamma function) has nice properties and is convenient for computations. Denote this function by $\psi(x) = \frac{d}{dx} \log \Gamma(x) = \Gamma'(x)/\Gamma(x)$. Then, by differentiating the objective function, we find that the solution to the problem \eqref{s_MLE_weights} is given by:
$$
\psi(1+s-n) - \psi \left(1 + s - \frac{n}{2}\right) = \frac{1}{N \sum_{i=1}^N w_i} \tilde H(\hat a).
$$
For the particular case $n=3$ one can exploit properties of the digamma function \cite[Section 6.3]{Abramowitz} to recast the above equation as follows
$$
\sum_{k=1}^\infty \frac{1}{s-2+k} - \sum_{k=1}^\infty \frac{1}{s-\frac{1}{2}+k} = \frac{1}{N \sum_{i=1}^N w_i} \tilde H(\hat a).
$$
The last equation has been used for the step iii)-2 of experiments on the data in $\mathbb{B}^3$ reported in the next subsection.
\end{remark}

\subsection{Experiments}

In this subsection we present results in the three-dimensional hyperbolic ball.

\subsubsection{Experiment B1}

\noindent{\it Generation of the data set}

Data are generated from the following mixture of three M\" obius distributions in $\mathbb{B}^3$:
\begin{enumerate}
\item[a)] The first cluster: $\pi_1=0.3$, $a_1=\{0.8,0,0\}$, $s_1=5$;

\item[b)] The second cluster: $\pi_2 =0.3$, $a_2=\{0,0.8,0\}$, $s_2=4$;

\item[c)] The third cluster:  $\pi_3=0.4$, $a_3=\{0,0,0.8\}$, $s_3=3$.
\end{enumerate}

The sampling procedure is exposed in subsection \ref{sampling}. For the particular dimension $n=3$ \eqref{random_generate} reduces to polynomial equations:
$$
s^* = 3 \implies b^3=\kappa; \quad s^*=4 \implies \frac{15}{2} \left( \frac{b^3}{3}-\frac{b^5}{5}\right) = \kappa;
$$
$$
 s^*=5 \implies \frac{105}{2} \left( \frac{b^7}{7} - 2 \frac{b^5}{5} + \frac{b^3}{3}\right) = \kappa.
$$
The above formulae are used for generation of the data in this experiment, as well as in the next one.

\noindent{\it Results}

K-means found the following barycenters:
\begin{align*}
a_1&=\{0.840602, -0.0144454, -0.0103716\};\\
a_2&=\{0.00658405, 0.795361,0.00578596\};\\
a_3&=\{-0.0477645, -0.00233238, 0.780643\}.
\end{align*}

EM found the following mixture:
\begin{enumerate}
\item[a)] The first cluster: $\pi_1=0.299989$,
$a_1=\{0.840454, -0.0144339, -0.0103568\}$, $s_1=4.72472$;

\item[b)] The second cluster: $\pi_2 =0.30152$, $a_2=\{0.00632912,0.791876, -0.0222038\}$, $s_2=5.90229$;

\item[c)] The third cluster: $\pi_3=0.398491$, $a_3=\{-0.0439142, -0.0317803, 0.797197\}$, $s_3=3.17382$.
\end{enumerate}

\begin{figure*}[t]
\centering
  \begin{tabular}{@{}ccc@{}}
    \includegraphics[width=.3\textwidth]{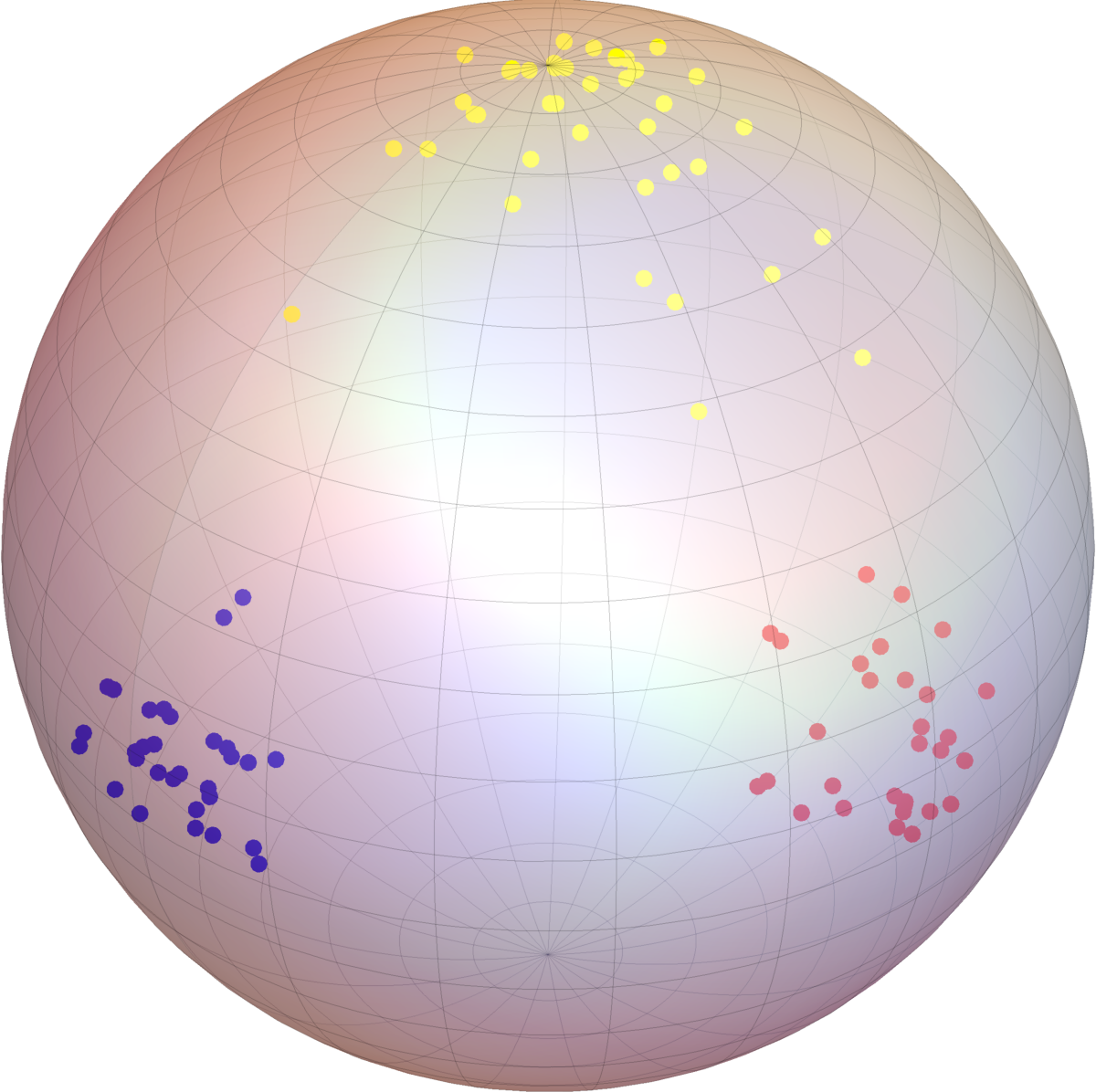}&\includegraphics[width=.3\textwidth]{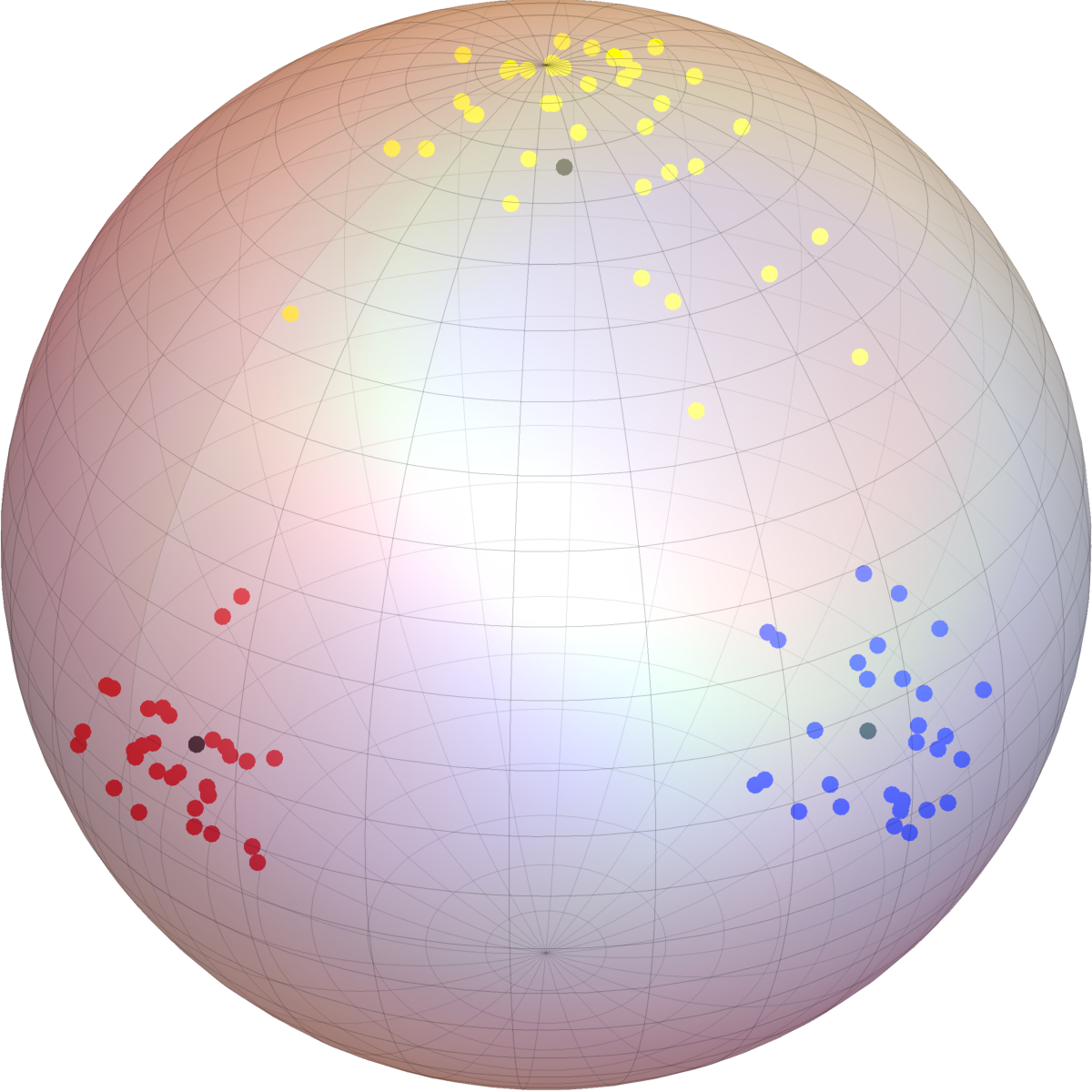}&\includegraphics[width=.3\textwidth]{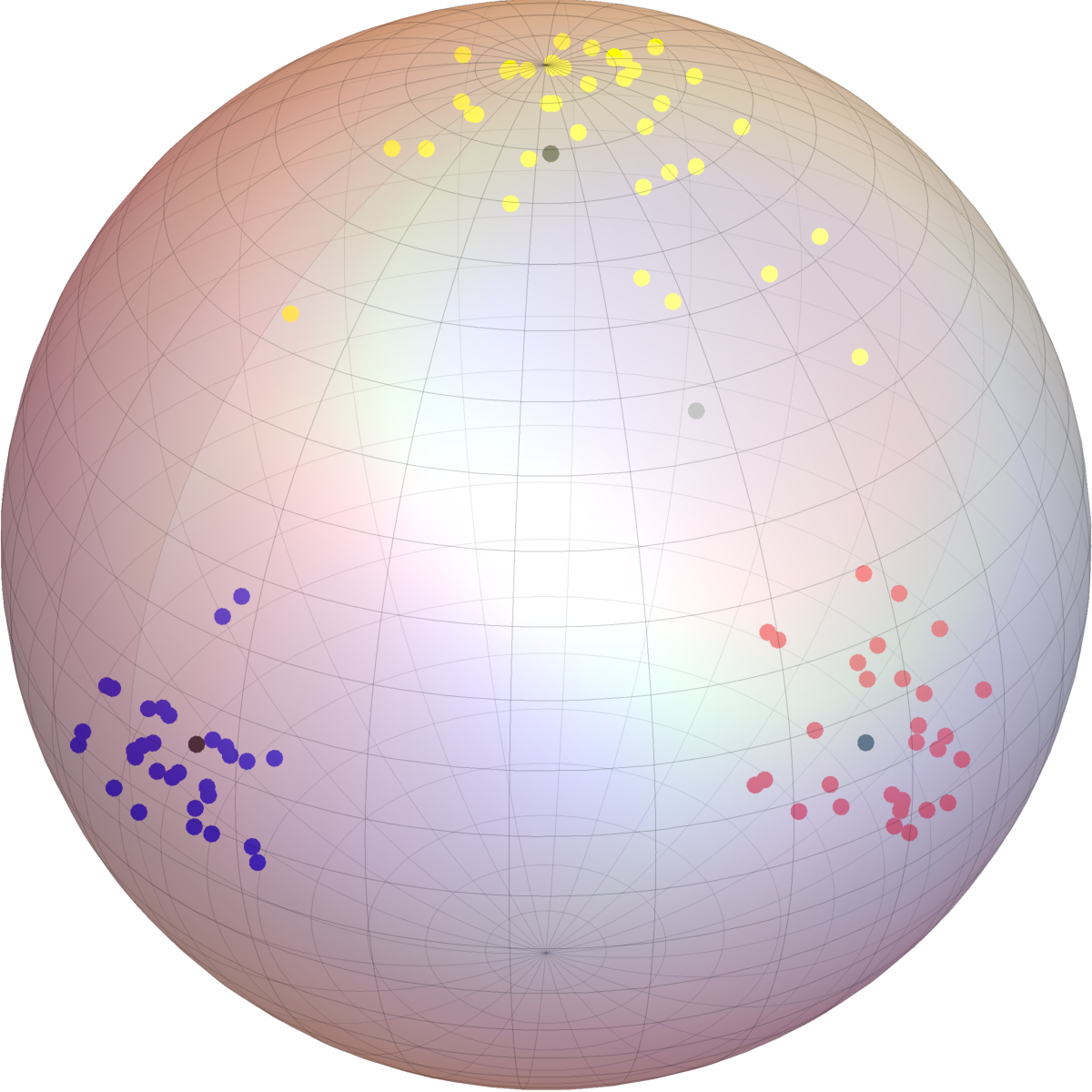}\\
    a) & b)  &c) \
  \end{tabular}
  \caption{\label{fig:10}Experiment B1: a) ground truth; b) k-means; c) EM algorithm}
\end{figure*}

\begin{figure*}[t]
\centering
  \begin{tabular}{@{}c@{}}
    \includegraphics[width=.4\textwidth]{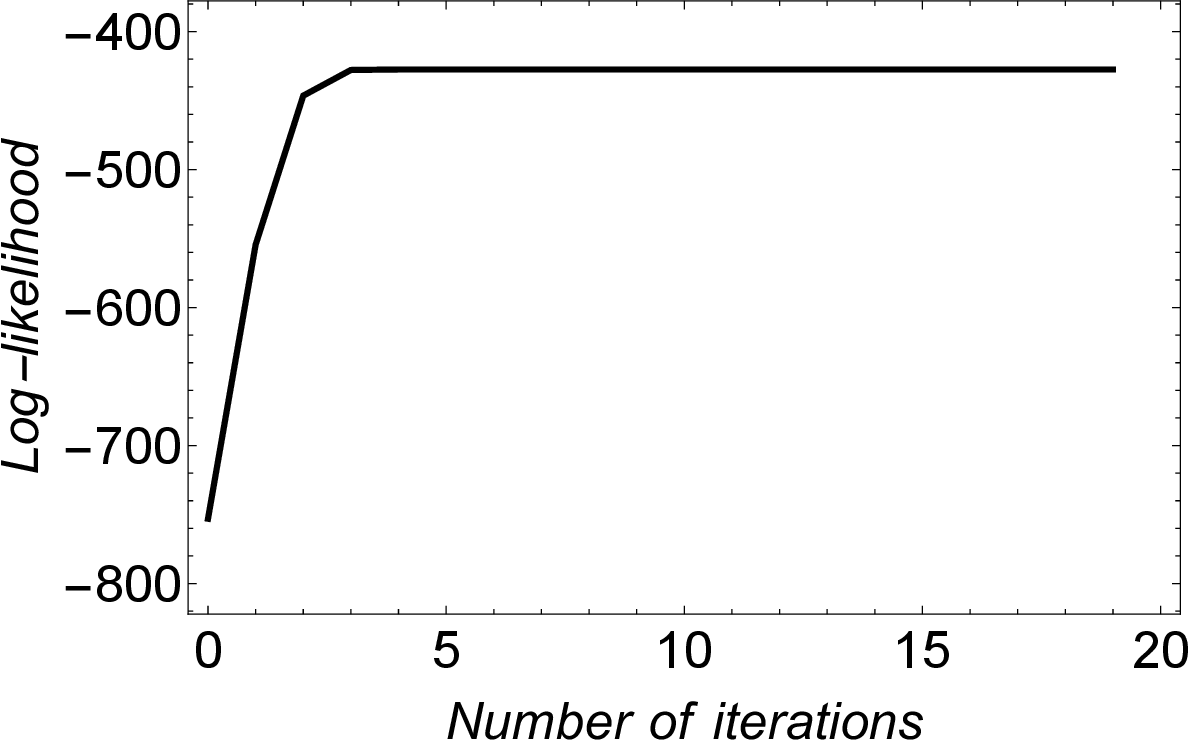}
  \end{tabular}
  \caption{\label{fig:11}Log-likelihood for EM in Experiment B1}
\end{figure*}

Results of the clustering are depicted in Figure \ref{fig:10} with the evolution of the log-likelihood function for the EM algorithm shown in Figure \ref{fig:11}. Notice one gray point in Figure \ref{fig:10}c) demonstrating that one observation has not been decisively assigned to any cluster.

\subsubsection{Experiment B2}

\noindent{\it Generation of the data set}

Data are generated from the mixture of four M\" obius distributions in $\mathbb{B}^3$:
\begin{enumerate}
\item[a)] $\pi_1=0.35$, $a_1=\{0, -0.4, 0.4\}$, $s_1=5$;

\item[b)] $\pi_2 =0.25$, $a_2=\{-0.4, -0.2, -0.3\}$, $s_2=5$;

\item[c)] $\pi_3=0.2$, $a_3=\{0.6,-0.6,0\}$, $s_3=4$;

\item[d)] $\pi_4=0.2$, $a_4=\{0, 0, 0.85\}$, $s_4=3$.
\end{enumerate}

\noindent{\it Results}

K-means found the following barycenters:
\begin{align*}
a_1&=\{-0.00723196,-0.445543,0.401501\};\\
a_2&=\{-0.36037,-0.164732,-0.284476\};\\
a_3&=\{0.596384,-0.592481,0.0260412\};\\
a_4&=\{0.000648593,-0.0111959,0.857676\}.
\end{align*}

EM found the following mixture:
\begin{enumerate}
\item[a)] The first cluster: $\pi_1=0.348847$, $a_1=\{-0.00646996,-0.406079,0.382212\}$, $s_1=4.23968$;

\item[b)] The second cluster: $\pi_2 =0.252501$, $a_2=\{-0.362144,-0.17027,-0.294944\}$, $s_2=5.18885$;

\item[c)] The third cluster: $\pi_3=0.199843$, $a_3=\{0.601346,-0.604124,0.0337518\}$, $s_3=7.32962$;

\item[d)] The forth cluster: $\pi_4=0.198809$, $a_4=\{-0.00285491,-0.00712774,0.860007\}$, $s_4=2.88586$.
\end{enumerate}

\begin{figure*}[t]
\centering
  \begin{tabular}{@{}ccc@{}}
    \includegraphics[width=.3\textwidth]{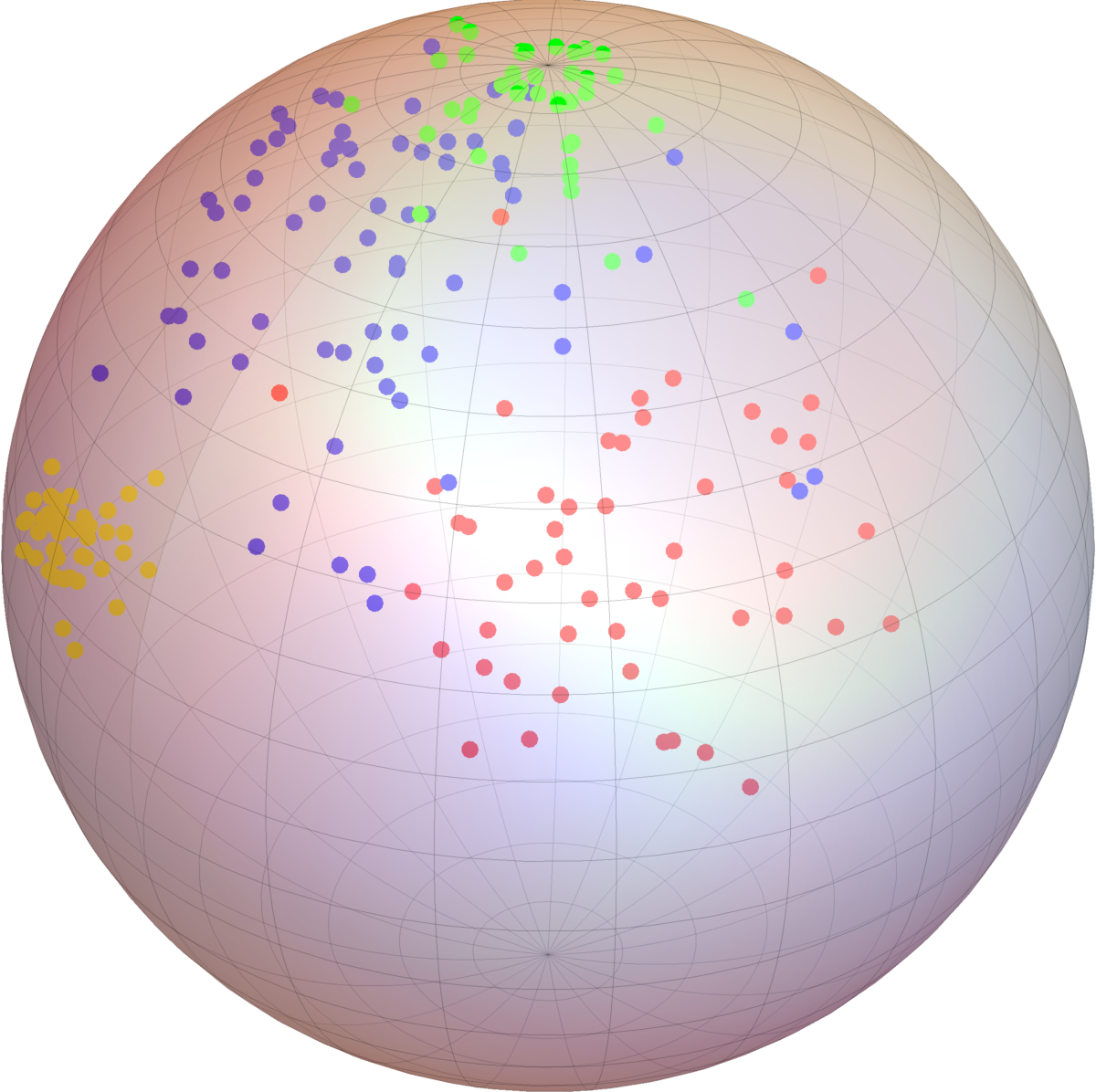}&\includegraphics[width=.3\textwidth]{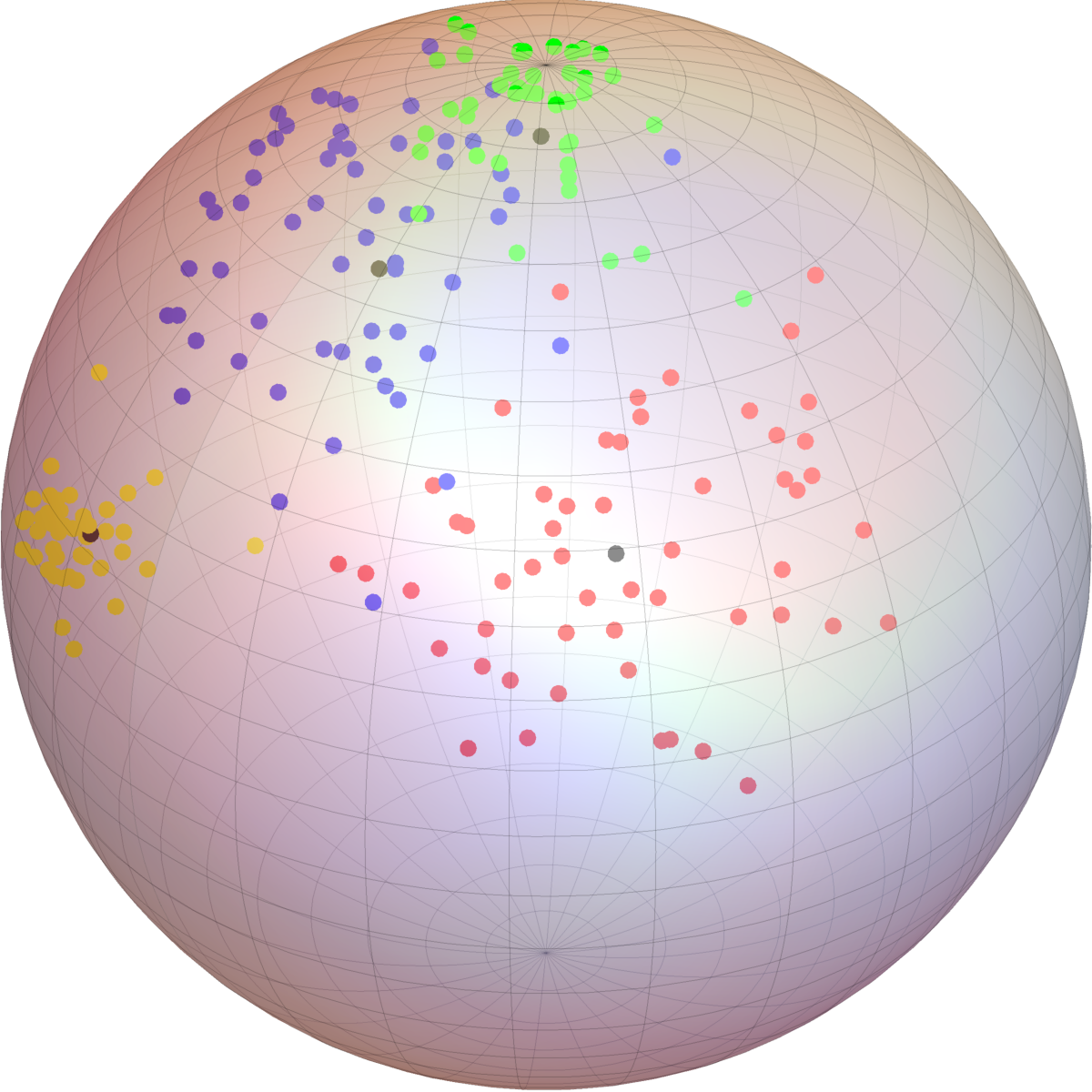}&\includegraphics[width=.3\textwidth]{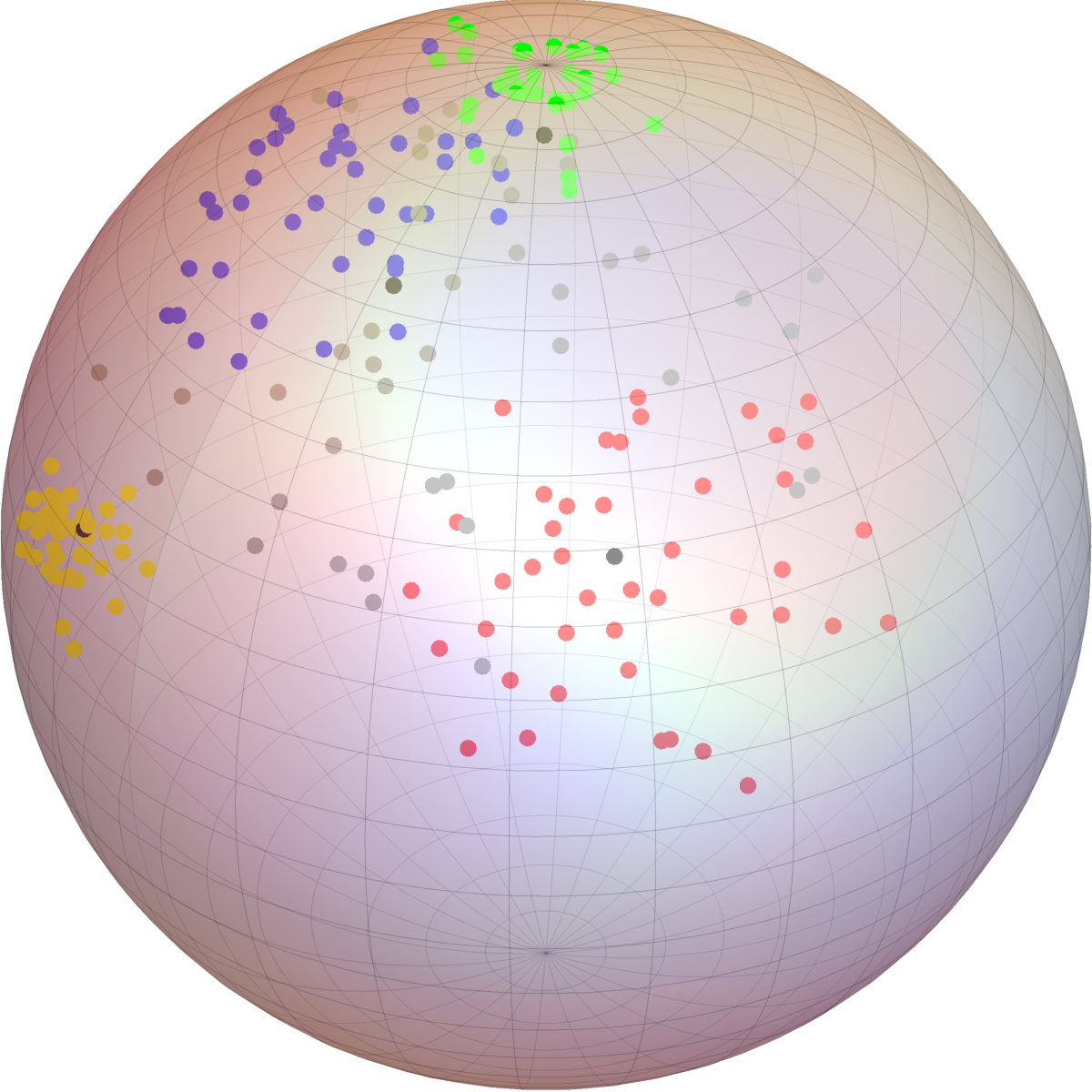}\\
    a) & b)  &c) \
  \end{tabular}
  \caption{\label{fig:12}Experiment B2: a) ground truth; b) k-means; c) EM algorithm}
\end{figure*}

\begin{figure*}[t]
\centering
  \begin{tabular}{@{}c@{}}
    \includegraphics[width=.4\textwidth]{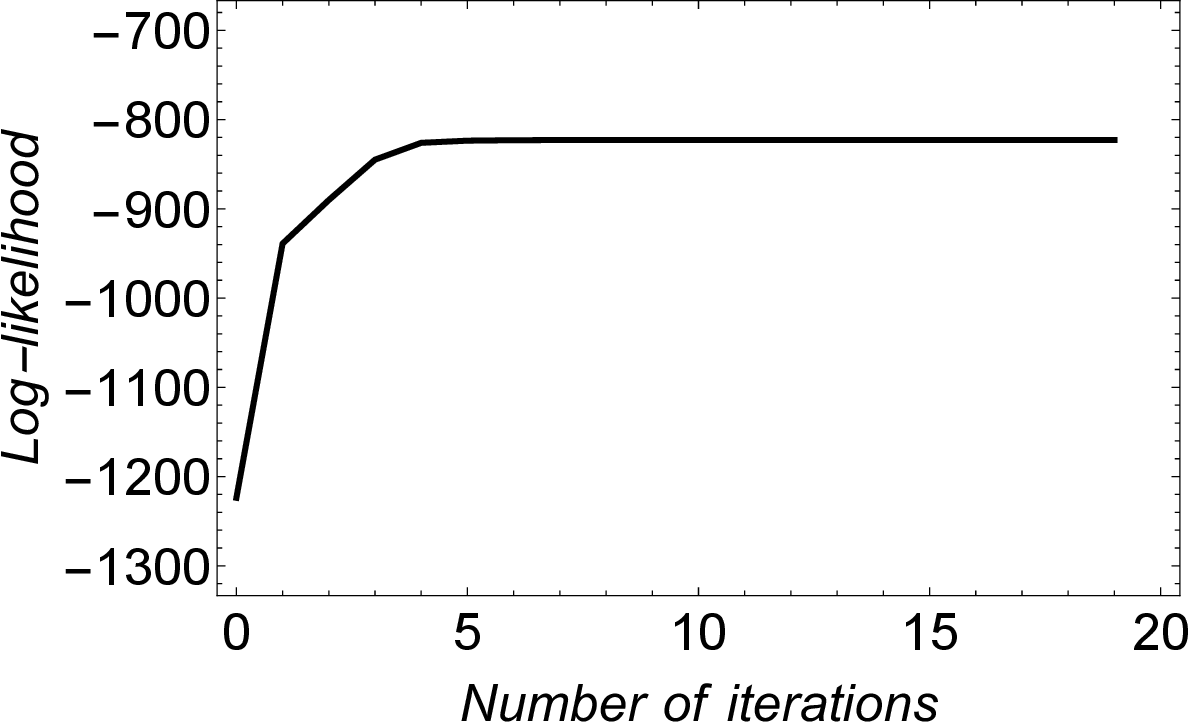}
  \end{tabular}
  \caption{\label{fig:13}Log-likelihood for EM in Experiment B2}
\end{figure*}

Results of Experiment B2 are visualized in Figure \ref{fig:12}. Evolution of the log-likelihood function for the EM algorithm is shown in Figure \ref{fig:13}. Notice a significant number of gray points in Figure \ref{fig:12}c).

\section{Conclusion and outlook} \label{sec:7}
The data never comes in the form of points in hyperbolic spaces. Therefore, hyperbolic ML requires demanding supervised learning algorithms for embedding the data. This motivated extensive research efforts which yielded principled approaches to this problem.

However, faithful data embedding is only the first phase of comprehensive ML solutions. Once the data are embedded into hyperbolic spaces there is the necessity for data analysis, statistical inference and generative modeling. In the present paper we have introduced two central algorithms of unsupervised learning in hyperbolic balls: $k$-means and learning mixture models via EM algorithm.

The proposed clustering algorithms are based on the novel definition of the mean (barycenter) in hyperbolic balls and the novel family of probability distributions in hyperbolic balls. These mathematical notions are conformally invariant, as they respect geometry of the balls. This fact renders them tractable and thus suitable for efficient computations. Most important, there is an efficient procedure for computation of the barycenter, as well as for maximum likelihood estimation of parameters. Notice, however, that computation of the barycenter in hyperbolic spaces is inevitably more demanding than computation of the average in Euclidean spaces. This is one of computational costs that must be paid when working with hyperbolic representations. On the other hand, most of the essential mathematical techniques (such as MLE) are not more demanding than their counterparts in Euclidean spaces.

Convergence of the proposed algorithms is not discussed here, since it can be established in an analogous way as for their counterparts in Euclidean spaces. Of course, just like in Euclidean spaces, there is no guarantee of finding the global optimum. In most cases methods will converge to local optima.

In conclusion, we believe that the present study enhances both theoretic and algorithmic foundations of hyperbolic ML. It extends the available machinery, thus contributing towards future hyperbolic DL pipelines.


\begin{thebibliography}{19}

\bibitem{Smith} Smith, A.L., Asta, D.M. \& Calder, C.A. The geometry of continuous latent space models for network data. {\em Statistical Science: A Review Journal of the Institute of Mathematical Statistics}. \textbf{34}, 428 (2019)

\bibitem{Sala} Sala, F., De Sa, C., Gu, A. \& Ré, C. Representation tradeoffs for hyperbolic embeddings. {\em Proceedings of the 35th International Conference on Machine Learning}. pp. 4460-4469 (2018)

\bibitem{Arvanitidis} Arvanitidis, G., Hansen, L.K. \& Hauberg, S. Latent Space Oddity: on the Curvature of Deep Generative Models. {\em International Conference on Learning Representations}. (2018)

\bibitem{Nickel} Nickel, M. \& Kiela, D. Poincaré embeddings for learning hierarchical representations. {\em Proceedings of the 31st International Conference on Neural Information Processing Systems}. pp. 6341-6350 (2017)

\bibitem{Tifrea} Tifrea, A., Bécigneul, G. \& Ganea, O.-E. Poincaré GloVe: Hyperbolic Word Embeddings. {\em International Conference on Learning Representations}. (2019)

\bibitem{Muscoloni} Muscoloni, A., Thomas, J.M., Ciucci, S., Bianconi, G. \& Cannistraci, C.V. Machine learning meets complex networks via coalescent embedding in the hyperbolic space. {\em Nature Communications}. \textbf{8}, 1615 (2017)

\bibitem{Chami} Chami, I., Wolf, A., Juan, D.-C., Sala, F., Ravi, S. \& Ré, C. Low-dimensional hyperbolic knowledge graph embeddings. {\em ArXiv:2005.00545}. (2020)

\bibitem{Corso} Corso, G., Ying, R., Pándy, M., Veli\v ckovi\' c, P., Leskovec, J. \& Lio, P. Neural distance embeddings for biological sequences. {\em Proceedings of the 35th International Conference on Neural Information Processing Systems}. pp. 18539-18551 (2021)

\bibitem{Facebook} Klimovskaia, A., Lopez-Paz, D., Bottou, L. \& Nickel, M. Poincaré maps for analyzing complex hierarchies in single-cell data. {\em Nature Communications}. \textbf{11}, 2966 (2020)

\bibitem{Li} Li, A., Yang, B., Huo, H., Chen, H., Xu, G. \& Wang, Z. Hyperbolic neural collaborative recommender. {\em IEEE Transactions on Knowledge and Data Engineering}. \textbf{35}, 9114-9127 (2022)

\bibitem{Chamberlain} Chamberlain, B.P., Hardwick, S.R., Wardrope, D.R., Dzogang, F., Daolio, F. \& Vargas, S. Scalable hyperbolic recommender systems. {\em ArXiv:1902.08648}. (2019)

\bibitem{Baker} Baker, C., Suárez-Méndez, I., Smith, G., Marsh, E.B., Funke, M., Mosher, J.C., Maestú, F., Xu, M. \& Pantazis, D. Hyperbolic graph embedding of MEG brain networks to study brain alterations in individuals with subjective cognitive decline. {\em IEEE Journal of Biomedical and Health Informatics}. \textbf{28}, 7357-7368 (2024)

\bibitem{Allard} Allard, A. \& Serrano, M.A. Navigable maps of structural brain networks across species. {\em PLoS Computational Biology}. \textbf{16}, e1007584 (2020)

\bibitem{Dempster} Dempster, A.P., Laird, N.M. \& Rubin, D.B. Maximum likelihood from incomplete data via the EM algorithm. {\em Journal of the Royal Statistical Society: Series B}. \textbf{39}, 1-22 (1977)

\bibitem{Banerjee} Banerjee, A., Dhillon, I.S., Ghosh, J. \& Sra, S. Clustering on the unit hypersphere using von Mises-Fisher distributions. {\em Journal Of Machine Learning Research}. \textbf{6}, 1345-1382 (2005)

\bibitem{Hamelryck} Mardia, K.V., Barber, S., Burdett, P.M., Kent, J.T. \& Hamelryck, T. Mixture models for spherical data with applications to protein bioinformatics. {\em Directional Statistics For Innovative Applications. Forum for Interdisciplinary Mathematics}. pp. 15-32 (2022)

\bibitem{JK} Ja\' cimovi\' c, V. \& Kalaj, D. Conformal and holomorphic barycenters in hyperbolic balls. {\em ArXiv:2410.02257}. (2024)

\bibitem{Jacimovic} Ja\' cimovi\' c, V. A group-theoretic framework for machine learning in hyperbolic spaces. {\em ArXiv:2501.06934}. (2025)

\bibitem{Abramowitz} Abramowitz, M. \& Stegun, I.A. Handbook of mathematical functions with formulas, graphs, and mathematical tables. (U.S. Government Printing Office, 1972)

\end{thebibliography}





\end{document}